%% file: arxiv.tex
\documentclass{article} 
\usepackage{iclr2026_conference,times}

\input{math_commands.tex}

\usepackage[section]{placeins}

\usepackage{hyperref}
\usepackage{url}
\usepackage{graphicx}
\usepackage{booktabs}
\usepackage{multirow}
\usepackage{amsmath,amssymb,amsfonts}
\usepackage{algorithmic}
\usepackage{enumitem}
\usepackage{float}
\usepackage{xcolor}
\usepackage{xspace}

\definecolor{kleinblue}{RGB}{0,47,167}
\definecolor{rqone}{RGB}{37,99,235}
\definecolor{rqtwo}{RGB}{217,119,6}
\definecolor{rqthree}{RGB}{5,150,105}
\hypersetup{colorlinks=true, citecolor=kleinblue, linkcolor=kleinblue, urlcolor=kleinblue}

\newcommand{\qbplus}{\texttt{QuanBench+}\xspace}
\newcommand{\RQone}{\textcolor{rqone}{\textbf{RQ1}}}
\newcommand{\RQtwo}{\textcolor{rqtwo}{\textbf{RQ2}}}
\newcommand{\RQthree}{\textcolor{rqthree}{\textbf{RQ3}}}
\newcommand{\Aone}{\textcolor{rqone}{\textbf{A1}}}
\newcommand{\Atwo}{\textcolor{rqtwo}{\textbf{A2}}}
\newcommand{\Athree}{\textcolor{rqthree}{\textbf{A3}}}

\graphicspath{{figures/}}

\title{\qbplus\\A Unified Multi-Framework Benchmark\\for LLM-Based Quantum Code Generation}

\author{
{\bfseries Ali Slim}$^{1}$\thanks{The first four authors contributed equally to this work.} \quad
{\bfseries Haydar Hamieh}$^{1}$\footnotemark[1] \quad
{\bfseries Jawad Kotaich}$^{1}$\footnotemark[1] \quad
{\bfseries Yehya Ghosn}$^{1}$\footnotemark[1] \\
{\bfseries Mahdi Chehimi}$^{1}$ \quad
{\bfseries Ammar Mohanna}$^{1}$ \quad
{\bfseries Hasan Abed Al Kader Hammoud}$^{2}$ \quad
{\bfseries Bernard Ghanem}$^{2}$ \\
\\[0.35em]
\normalfont $^{1}$ American University of Beirut \\
\normalfont $^{2}$ King Abdullah University of Science and Technology
}

\iclrfinalcopy 

\begin{document}

\maketitle

\begin{abstract}
Large Language Models (LLMs) are increasingly used for code generation, yet quantum code generation is still evaluated mostly within single frameworks, making it difficult to separate quantum reasoning from framework familiarity. We introduce \qbplus, a unified benchmark spanning Qiskit, PennyLane, and Cirq, with 42 aligned tasks covering quantum algorithms, gate decomposition, and state preparation.

We evaluate models with executable functional tests, report Pass@1 and Pass@5, and use KL-divergence-based acceptance for probabilistic outputs. We additionally study Pass@1 after feedback-based repair, where a model may revise code after a runtime error or wrong answer. Across frameworks, the strongest one-shot scores reach 59.5\% in Qiskit, 54.8\% in Cirq, and 42.9\% in PennyLane; with feedback-based repair, the best scores rise to 83.3\%, 76.2\%, and 66.7\%, respectively. These results show clear progress, but also that reliable multi-framework quantum code generation remains unsolved and still depends strongly on framework-specific knowledge. \\[3pt]
\textbf{Keywords:} large language models, quantum programming, benchmarking,\\
Qiskit, PennyLane, Cirq
\end{abstract}

\section{Introduction}
LLMs have achieved strong performance on classical code-generation benchmarks such as HumanEval~\cite{humaneval} and related variants. As quantum computing moves further into software practice, developers increasingly rely on ecosystems such as Qiskit~\cite{qiskit}, PennyLane~\cite{pennylane}, and Cirq~\cite{cirq}. The practical question is no longer whether models can emit quantum-flavored code, but whether they can generate \emph{correct} quantum programs across frameworks with different abstractions and APIs.

Quantum programming differs from classical programming in that program outputs are typically \emph{probabilistic} measurement statistics rather than deterministic values. A qubit is represented as $\lvert \psi\rangle = \alpha\lvert 0\rangle + \beta\lvert 1\rangle$, where $\lvert\alpha\rvert^2$ and $\lvert\beta\rvert^2$ denote measurement probabilities \cite{QuantumBook}. As a result, correctness must be defined in terms of output distributions, measurement schemes, and execution settings.

Several quantum-code benchmarks have emerged, including Qiskit HumanEval~\cite{QiskitHumanEval}, QHackBench~\cite{QHackBench}, QCircuitBench~\cite{QCircuitBench}, and QuanBench~\cite{QuanBench}. Many prior evaluations are still tied to a single framework or do not explicitly hold task intent fixed across frameworks, making it hard to disentangle quantum reasoning from familiarity with the framework.

A multi-framework benchmark is valuable because it exposes two distinct failure modes: (i) \emph{conceptual} errors in quantum reasoning (e.g., incorrect algorithmic structure or measurement logic) and (ii) \emph{framework} errors (e.g., wrong APIs, missing measurements, simulator misuse). Without controlling the task intent across frameworks, it is hard to attribute failures to one category or the other.

We therefore introduce \qbplus, a unified multi-framework evaluation that holds task intent constant while varying only the target framework.

\paragraph{Research questions (RQs).}
We organize the paper around the following questions:
\begin{itemize}[leftmargin=*, itemsep=0.25em, topsep=0.25em]
    \item \RQone: How accurately can modern LLMs generate \emph{correct} quantum code across Qiskit, PennyLane, and Cirq?
    \item \RQtwo: To what extent are observed gains driven by framework-specific boilerplate (prefill) versus true task-level reasoning?
    \item \RQthree: How much can an automated feedback loop improve one-shot performance under the same functional test harness?
\end{itemize}

\paragraph{Answers in brief (A1--A3).}
Our experiments answer these questions as follows:
\begin{itemize}[leftmargin=*, itemsep=0.25em, topsep=0.25em]
    \item \Aone: Current models show real progress, but cross-framework reliability remains low and strongly framework-dependent.
    \item \Atwo: Prefill mainly reduces interface friction and boilerplate mistakes; it does not remove the harder semantic failures.
    \item \Athree: Feedback-based repair recovers a substantial share of first-attempt failures, but the remaining errors are still dominated by reasoning mistakes.
\end{itemize}

\paragraph{Contributions.}
This paper makes the following contributions:
\begin{itemize}
    \item We introduce \qbplus, a unified multi-framework benchmark spanning Qiskit, PennyLane, and Cirq.
    \item We adapt 42 tasks into framework-aligned prompts that preserve the same functional goal across ecosystems and support automated grading.
    \item We standardize evaluation with executable Pass@k testing and KL-divergence-based acceptance for probabilistic outputs, and report Pass@1, Pass@5, and Pass@1 after feedback-based repair.
    \item We characterize where performance changes come from by comparing frameworks, prefill conditions, error types, and iterative repair.
\end{itemize}

\paragraph{Paper organization.}
The remainder of the paper is organized as follows. Section \ref{sec:related_work} reviews related works, and Section~\ref{sec:metrics} defines the considered evaluation criteria. Then, Sections \ref{sec:benchmark} and \ref{sec:experiments} describe the benchmark model and experimental setup, respectively. Next, Section \ref{sec:results} presents the main results, and Sections \ref{sec:discussion} and \ref{sec:conclusion} close with discussion and conclusions.

\section{Related Work}\label{sec:related_work}
\subsection{General Code Generation Benchmarks}
HumanEval~\cite{humaneval} and HumanEval+~\cite{humanevalplus} established executable functional evaluation as a standard way to assess LLM code generation. Their success made Pass@k-style testing and fixed harnesses the default for classical code, but their deterministic task design does not transfer cleanly to probabilistic quantum programs.

\subsection{Quantum Code Generation Benchmarks}
A growing body of work evaluates LLMs for quantum programming. Qiskit HumanEval~\cite{QiskitHumanEval} measures proficiency with the Qiskit API, QHackBench~\cite{QHackBench} focuses on PennyLane tasks derived from QHack challenges, QCircuitBench~\cite{QCircuitBench} targets larger-scale circuit generation, and QuanBench~\cite{QuanBench} curates tasks spanning algorithms, state preparation, and decomposition. QCoder Benchmark~\cite{mikuriya2025qcoder} further connects generation to execution by incorporating simulator-based feedback.

Related work also targets domain-specific assistants and training resources. Qiskit Code Assistant~\cite{dupuis2024qiskit} and subsequent work on quantum verifiable rewards~\cite{dupuis2025quantum} study specialized Qiskit generation, while Pennylang~\cite{basit2025pennylang} and PennyCoder~\cite{basit2025pennycoder} focus on the PennyLane ecosystem. QUASAR extends the problem toward tool-augmented quantum assembly generation~\cite{yu2025quasar}. The common limitation is scope: most evaluations remain tied to a single framework or a single layer of the tooling stack.

\subsection{Positioning of \qbplus}
\qbplus extends this line of work by holding task objectives fixed while varying the target framework. That design makes it possible to ask a more useful question: whether observed performance reflects portable quantum reasoning or simply better recall of one framework's conventions.

\section{Evaluating Quantum Code Generation}\label{sec:metrics}
We follow the functional-correctness paradigm used in HumanEval~\cite{humaneval}: a generated program is considered correct if it executes and satisfies a task-specific correctness criterion under a fixed harness. In our setting, tasks either admit deterministic checks or require distributional agreement of measurement outcomes.

\subsection{Correctness Metrics}
\paragraph{Pass@k.}
We use Pass@k as our primary correctness metric. Pass@k measures the probability that at least one of the top-$k$ generated solutions is correct:
\begin{equation}
\mathrm{Pass@}k = 1 - \frac{\binom{n-c}{k}}{\binom{n}{k}},
\label{eq:passk}
\end{equation}
where $n$ is the number of generated samples and $c$ is the number of correct samples. We report Pass@1 and Pass@5 in this version.

\paragraph{KL divergence for probabilistic outputs.}
We compute the KL divergence between the canonical distribution $P$ and the model-generated distribution $Q$:
\begin{equation}
D_{\mathrm{KL}}(P \Vert Q) = \sum_x P(x)\,\log\frac{P(x)}{Q(x)}.
\label{eq:kl}
\end{equation}
To avoid undefined values when $Q(x)=0$ for states with $P(x)>0$, we apply a small additive smoothing constant $\varepsilon$ to both distributions before renormalization. A solution is accepted when the resulting divergence is below a global threshold set to $0.05$. Appendix~\ref{sec:kl_threshold_proof} calibrates this threshold from repeated canonical executions of the probabilistic tasks.

\subsection{Why We Exclude Fidelity}
QuanBench~\cite{QuanBench} additionally reports a \emph{process fidelity} (unitary overlap) between a reference circuit and a generated circuit,
\begin{equation}
\mathcal{F}(U_{\mathrm{ref}}, U_{\mathrm{gen}})=\left|\frac{1}{d}\operatorname{Tr}\left(U_{\mathrm{ref}}^\dagger U_{\mathrm{gen}}\right)\right|^2,\qquad d=2^{n_q},
\end{equation}
where $n_q$ is the number of qubits.
It measures global similarity at the level of the implemented unitary.

In \qbplus, we define correctness operationally as \emph{task success}: a solution is correct if it produces the required measurement statistics (or output probability distribution) under the prompt-specified inputs and measurement scheme. Under this definition, many circuits can be \emph{task-equivalent} while having low unitary-overlap fidelity. For example, inserting basis-dependent phase transformations can preserve computational-basis measurement probabilities for a task while changing $\mathcal{F}(U_{\mathrm{ref}}, U_{\mathrm{gen}})$ substantially.

More generally, compilation and optimization routinely transform circuits into syntactically different realizations that are functionally equivalent. This motivates a large body of work on \emph{quantum circuit equivalence checking}, which explicitly targets the question ``do two differently structured circuits implement the same functionality?'' using decision-diagram representations, reversible miters, and SAT-/simulation-based techniques~\cite{burgholzer2021advanced,yamashita2010fast}. In our benchmark setting, fidelity can therefore yield \emph{false negatives}: penalizing prompt-correct solutions that differ globally from a canonical reference circuit but still solve the task.

Additionally, fidelity/infidelity is an average-case notion and may not align with task-relevant error, particularly under coherent noise: relationships between experimentally reported average error rates (or infidelity-like quantities) and worst-case measures can differ by orders of magnitude~\cite{kueng2016comparing,wallman2015bounding}. Since our goal is to benchmark prompt-level functional correctness across frameworks, we prioritize executable functional evaluation (Pass@k) and distributional comparison (KL divergence) as primary scoring criteria, and disregard fidelity as a correctness metric.
\section{\qbplus Benchmark}\label{sec:benchmark}

\subsection{Benchmarking Workflow}
We follow a standard benchmarking workflow: define the objective, choose metrics aligned with task outputs, control the execution environment, construct paired prompts and canonical solutions, select representative models and frameworks, and assess outputs under one automated harness.

\begin{figure*}[htbp]
\centering
\includegraphics[width=\linewidth]{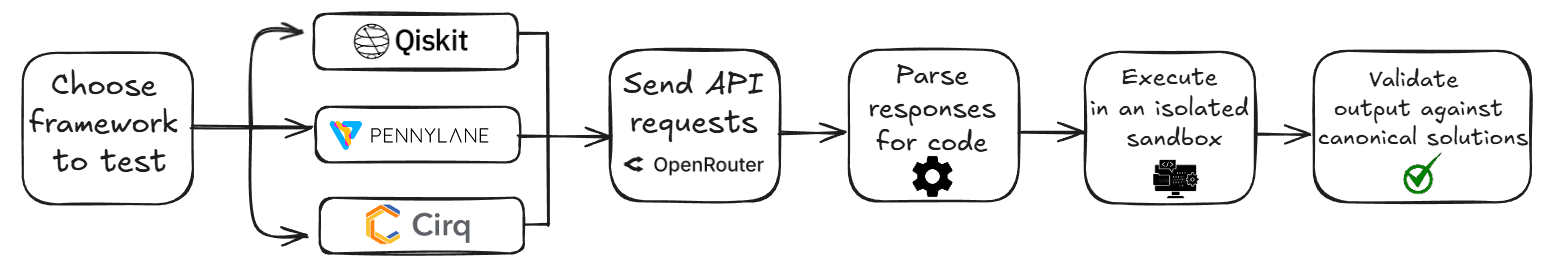}
\caption{\textbf{The benchmark holds task intent and execution conditions fixed across frameworks.} Our workflow standardizes prompts, grading, and runtime settings before comparing models on Qiskit, PennyLane, and Cirq.}
\label{fig:benchmark_pipeline}
\end{figure*}

\subsection{Task Set and Categories}
\qbplus is derived from the original QuanBench task set~\cite{QuanBench}. We retain tasks that admit clear numerical or functional correctness criteria and adapt them to Qiskit, PennyLane, and Cirq while preserving their objectives. Prompts were modified to account for framework-specific APIs and library conventions. Two tasks from the original benchmark were removed because they did not support reliable cross-framework grading. The final benchmark contains 42 tasks spanning three categories:
\begin{itemize}
    \item \textbf{Quantum Algorithms}
    \item \textbf{Gate Decomposition}
    \item \textbf{State Preparation}
\end{itemize}

\subsection{Prompt Standardization and Output Normalization}
To ensure fair comparisons, the set of canonical solutions is unified for all models across all frameworks. Each model receives the same prompt per task and framework with strict instructions on code-only output and expected function interfaces. For tasks requiring inputs, a random set of non-trivial inputs was generated once and used across all models and frameworks. Each canonical solution's output is standardized to a probability array representing the measurement distribution over computational basis states.

\subsection{Modifications on Prompts and Canonical Solutions}
\paragraph{Prompt Modifications.}
All prompts were modified to ensure that the correct libraries were imported for each framework. In addition, we enforced that models return code only, without any accompanying explanation, to improve execution efficiency. This requirement was explicitly stated at the beginning of each prompt.

\FloatBarrier
\begin{table}[!htbp]
\caption{\textbf{Only a small subset of tasks required benchmark-level edits.} These prompt changes and removals were needed to make grading consistent across Qiskit, PennyLane, and Cirq.}
\label{tab:dropped_tasks}
\centering
\begin{tabular}{c p{3.2cm} p{6.8cm}}
\toprule
Task & Edit & Rationale \\
\midrule
5  & Task removed & Did not output a clear probabilistic answer, making quantitative evaluation unreliable. \\
25 & Prompt modified & The old version measures all qubits, while the new version measures the first 3 qubits. \\
28 & Prompt modified & The prompt did not clearly instruct measuring all qubits, leading to ambiguous output distributions. \\
38 & Task removed & The testing procedure for the machine learning task was not clearly specified, preventing consistent evaluation. \\
41 & Prompt modified & The randomized input library was replaced by a pre-decided randomly generated input. \\
\bottomrule
\end{tabular}
\end{table}
\FloatBarrier

\section{Experimental Setup}\label{sec:experiments}

\subsection{Models}
We evaluate a diverse set of frontier and open-weight LLMs (listed in Appendix ~\ref{sec:app_models}), covering both models studied in QuanBench and more recent releases. All requests are issued through a unified API router. For Pass@1, we use greedy decoding (temperature $0.0$) and sample one completion per task. For Pass@5, we sample $k=5$ completions per task at temperature $0.8$.

\subsection{Execution Environment}
All generated solutions are executed in a controlled Python environment. To facilitate comparison with prior results, we use Python 3.10, Qiskit v0.46.0, Cirq v1.6.1, and PennyLane v0.43.1.

\paragraph{Execution and Grading Pipeline}
For each model completion, we apply the same evaluation procedure:
\begin{enumerate}[label=\textbf{P\arabic*:}, leftmargin=*, itemsep=0pt]
    \item Parse the completion and extract executable code.
    \item Execute the code in the target framework environment.
    \item Compare outputs using deterministic checks or a distributional threshold.
\end{enumerate}

\subsection{Feedback Loop}
In addition to standard one-shot generation, we evaluate a feedback loop setting that allows a model to repair its answer. The feedback loop is triggered on both runtime exceptions and wrong answer outputs. For each task, we execute the initial completion under the same harness used for Pass@k. If execution raises an exception (e.g., syntax/import/runtime errors), we provide the model with the exception trace and the original prompt, and request a corrected code-only solution. If the output of the generated code does not match the canonical solution, we provide the model with the wrong function and the original prompt, and request a corrected code-only solution. We report Pass@1 under this feedback loop as Pass@1 (FB). In all cases, we provide the models with a maximum of 5 repair chances.

\section{Results}\label{sec:results}
Three patterns dominate the results. Qiskit is consistently the easiest framework, PennyLane is consistently the hardest, and feedback-based repair recovers a large share of first-attempt failures without eliminating the remaining semantic mistakes. Detailed per-task maps and Pass@1-versus-Pass@5 comparisons are deferred to Appendices~\ref{sec:supp_pass1_pass5} and~\ref{sec:supp_heatmaps}.

\subsection{\RQone: Cross-Framework Functional Correctness}
\begin{figure}[htbp]
\centering
\includegraphics[width=\linewidth]{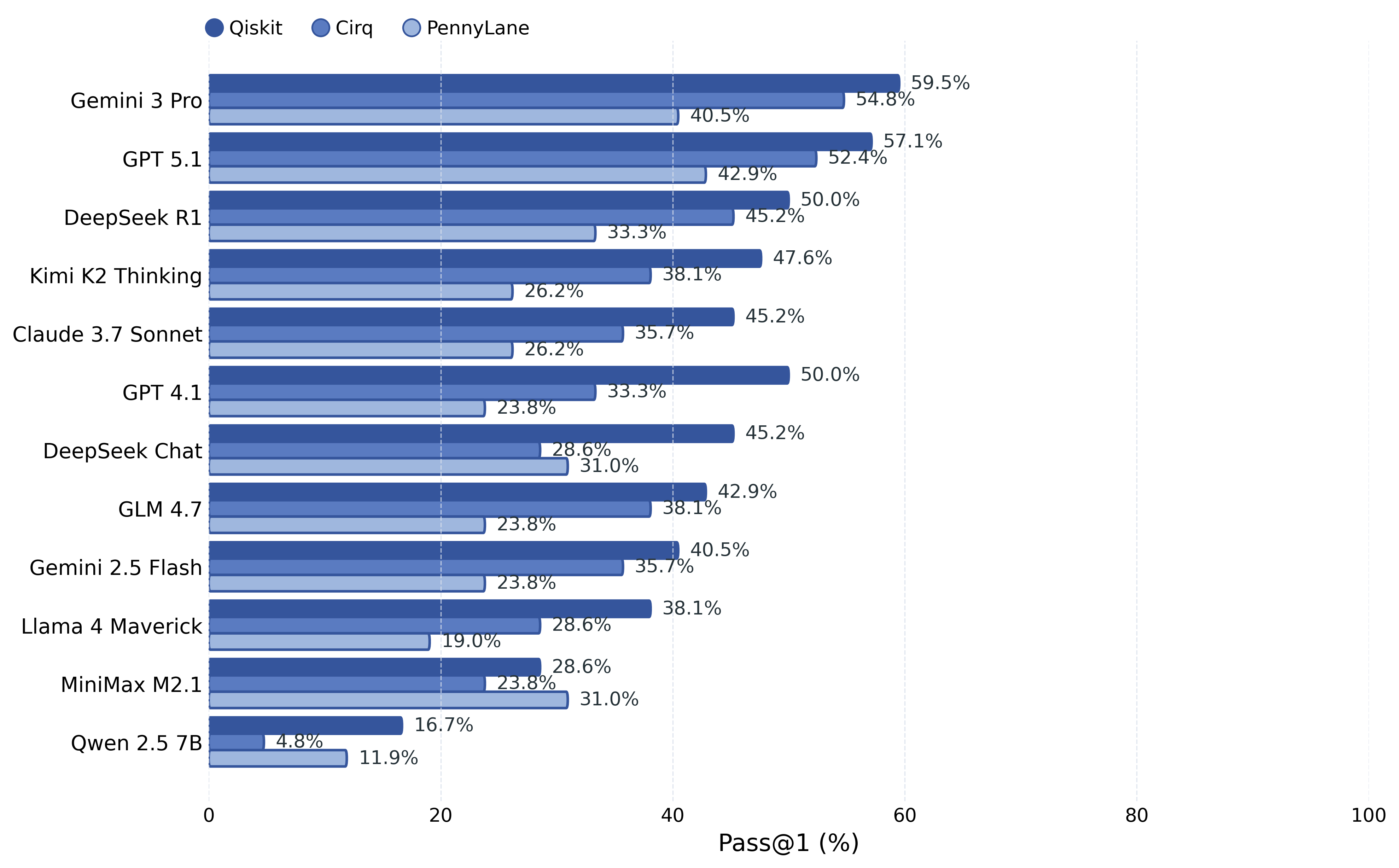}
\caption{\textbf{Qiskit is the easiest target and PennyLane the hardest under one-shot generation.} Models are ordered by average Pass@1 across frameworks, revealing both a stable ranking and a persistent framework gap.}
\label{fig:pass1_overview}
\end{figure}

Figure~\ref{fig:pass1_overview} provides the main one-shot ranking, while Appendix Table~\ref{tab:pass1_by_framework} reports the exact values. The strongest Pass@1 scores reach 59.5\% in Qiskit, 54.8\% in Cirq, and 42.9\% in PennyLane, which is enough to show real progress but not enough to claim dependable cross-framework generation.

The central finding is framework asymmetry rather than one universally dominant model. Gemini 3 Pro leads the average one-shot ranking because it is strongest on Qiskit and Cirq, whereas GPT-5.1 posts the best one-shot score on PennyLane. Across nearly every model, Qiskit sits highest and PennyLane lowest, indicating that framework-specific familiarity still explains a meaningful share of the variance.

\subsection{\RQtwo: Prefill vs No-Prefill}
Prefill mainly reduces interface friction rather than solving the hard reasoning cases. The appendix figures show that the largest gains tend to appear among smaller and mid-tier models, especially when framework boilerplate is easy to get wrong. Stronger models still benefit in some settings, but much less dramatically, which suggests that prefill helps most with imports, signatures, and setup rather than semantic program construction (Appendix~\ref{sec:supp_prefill_noprefill}).

\subsection{\RQthree: Feedback-Based Repair}
\begin{figure}[htbp]
\centering
\includegraphics[width=\linewidth]{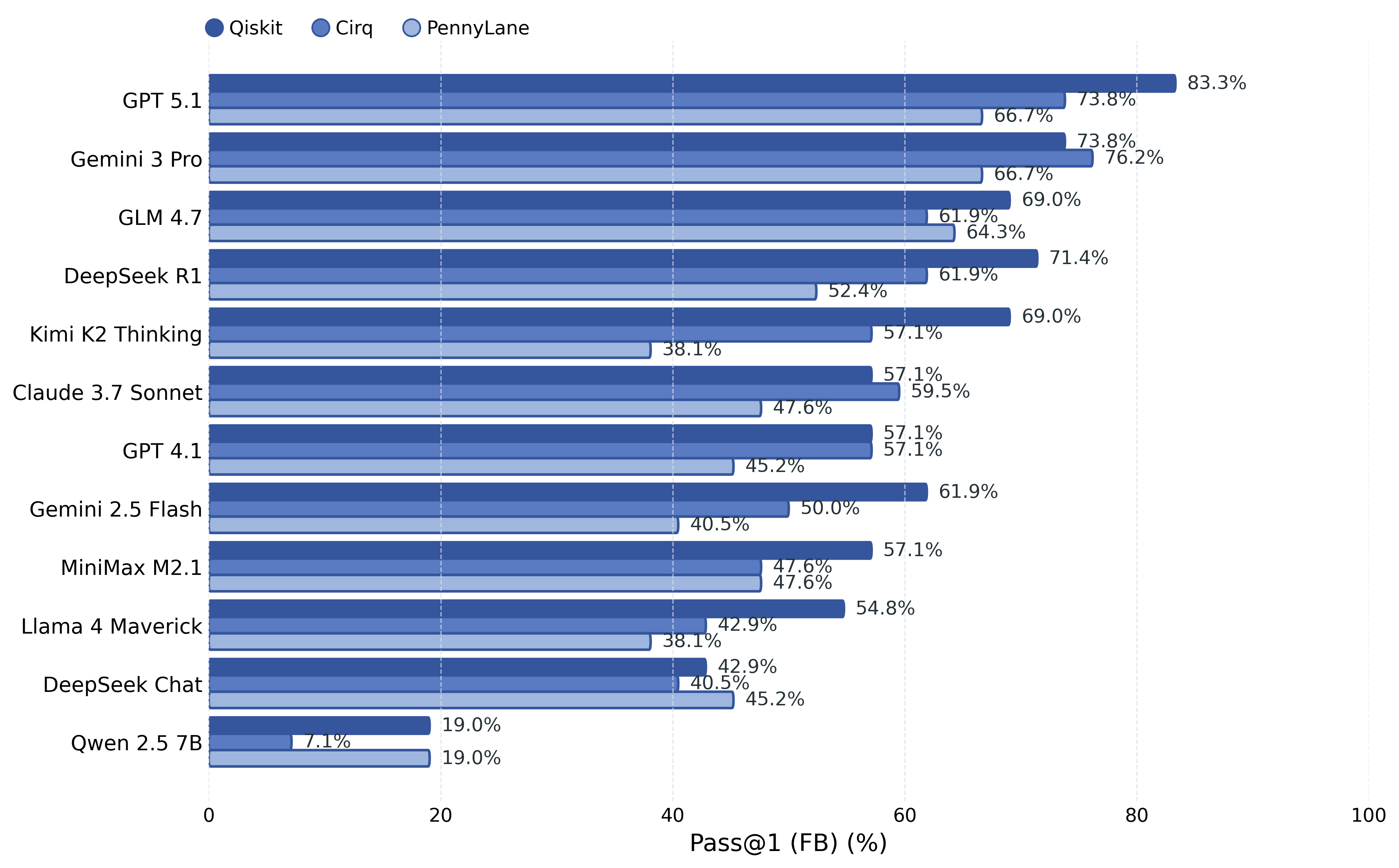}
\caption{\textbf{Feedback repair lifts accuracy across all three frameworks.} The gains are broad rather than model-specific, but no framework becomes fully reliable after repair.}
\label{fig:feedback_overview}
\end{figure}

Feedback-based repair materially lifts performance across all three frameworks. Figure~\ref{fig:feedback_overview} shows that the strongest repaired systems reach \textbf{83.3\%} in Qiskit, \textbf{76.2\%} in Cirq, and \textbf{66.7\%} in PennyLane. The gains are not limited to the frontier models: much of the middle of the ranking also improves sharply once runtime traces or wrong-answer signals are fed back to the model.

The improvement pattern matters as much as the headline numbers. Feedback narrows the gap caused by framework misuse and surface-level coding errors, but Appendix~\ref{sec:supp_feedback} shows that the remaining failures are still dominated by deeper semantic mistakes. Appendix Table~\ref{tab:pass1_by_framework} reports the exact Pass@1 and Pass@1 (FB) values used in the main paper.

\FloatBarrier
\FloatBarrier

We evaluate 42 tasks spanning quantum algorithms, state preparation, and gate decomposition; the task-count breakdown appears in Appendix~\ref{sec:app_task_categories}.

\section{Discussion}\label{sec:discussion}
The main result is not simply that newer models score higher; it is that difficulty remains strongly framework-dependent. Qiskit consistently yields the strongest outcomes, PennyLane remains harder even after repair, and Cirq typically falls in between. That pattern suggests current systems still rely heavily on framework-specific exposure and API familiarity rather than portable quantum programming competence.

We also observe a clear separation between errors that feedback can fix and errors that it cannot. Runtime and interface failures are often recoverable, but Appendix~\ref{sec:supp_error} and Appendix~\ref{sec:supp_feedback} show that the residual failures increasingly concentrate in deeper semantic mistakes.

\subsection{Threats to Validity}
Our evaluation depends on the correctness and completeness of canonical solutions. Cross-framework adaptation can introduce subtle mismatches between prompts and reference implementations, even when the intended task is the same. We mitigate this risk by excluding ambiguously graded tasks and reviewing framework-specific canonical code for functional equivalence.

A second threat is category imbalance. Quantum-algorithm tasks substantially outnumber state-preparation and decomposition tasks, which can amplify their influence on aggregate metrics and make the benchmark look harder wherever multi-step reasoning is required. Framework versioning is another source of uncertainty: a model may capture the right high-level intent while still failing execution because it reproduces stale APIs.

\subsection{Limitations \& Future Work}
\qbplus contains 42 tasks and therefore does not capture the full long tail of real-world quantum development. We also report only Pass@1, Pass@5, and Pass@1 (FB) in this version, which leaves out other potentially useful views of model behavior such as robustness to prompt variation, longer repair horizons, and tool-augmented workflows. Finally, the benchmark currently covers Qiskit, PennyLane, and Cirq; extending the same methodology to additional frameworks remains open future work.

\section{Conclusion}\label{sec:conclusion}
We answer \RQone, \RQtwo, and \RQthree{} by introducing \qbplus, a unified multi-framework benchmark for evaluating LLMs on quantum code generation in Qiskit, PennyLane, and Cirq. By adapting one task set across three ecosystems and grading outputs with executable functional tests, we provide a clearer picture of where current systems succeed, where they fail, and how much iterative repair can recover.

The headline conclusion is straightforward: modern models can often produce plausible quantum code, but reliable multi-framework correctness is still out of reach. Future progress will likely require more than model scale alone. It will depend on stronger exposure to quantum software data, better support for compositional reasoning and repair, and closer alignment with framework-specific APIs and execution patterns. We hope \qbplus provides a practical, reproducible basis for that next stage of evaluation\footnote{Source code: \url{https://github.com/JawadKotaichh/quanbench-plus}}.


\bibliographystyle{iclr2026_conference}
\bibliography{references}

\appendix
\include{arxiv_supplementary}

\end{document}

%% file: math_commands.tex
\usepackage{amsmath,amsfonts,bm}

\def\eqref#1{equation~\ref{#1}}

\def\1{\bm{1}}

\DeclareMathAlphabet{\mathsfit}{\encodingdefault}{\sfdefault}{m}{sl}
\SetMathAlphabet{\mathsfit}{bold}{\encodingdefault}{\sfdefault}{bx}{n}



%% file: arxiv_supplementary.tex
\section{Models Evaluated}
\label{sec:app_models}
The main paper focuses on comparative behavior; this appendix records the exact model list and the release references used to define the evaluated set.
\FloatBarrier
\begin{table}[H]
\caption{\textbf{The benchmark spans both frontier proprietary and open-weight systems.} This table lists the evaluated models and the release reference used for each one.}
\centering
\begin{tabular}{l l}
\toprule
\textbf{Model} & \textbf{Reference} \\
\midrule
\texttt{Claude-3.7-Sonnet} & \cite{claude37} \\
\texttt{DeepSeek-Chat} & \cite{deepseekv3} \\
\texttt{DeepSeek-R1} & \cite{deepseekv3} \\
\texttt{Gemini-2.5-Flash} & \cite{gemini25-flash} \\
\texttt{Gemini-3-Pro} & \cite{gemini3} \\
\texttt{GPT-4.1} & \cite{gpt4} \\
\texttt{GPT-5.1} & \cite{gpt5} \\
\texttt{Llama-4-Maverick} & \cite{llama4} \\
\texttt{Qwen-2.5-7B-Instruct} & \cite{qwen25} \\
\texttt{MiniMax-M2.1} & \cite{minimax} \\
\texttt{Z-ai-GLM-4.7} & \cite{zai} \\
\texttt{MoonshotAI-Kimi-K2-Thinking} & \cite{moonshot2024k2} \\
\bottomrule
\end{tabular}
\end{table}
\FloatBarrier

\section{Exact Main-Paper Result Tables}
\label{sec:app_exact_results}

The main paper emphasizes summary figures. This section records the exact one-shot and feedback-repair scores used in the core narrative.

\begin{table}[H]
\caption{\textbf{Feedback repair lifts scores across all three frameworks.} Exact Pass@1 and Pass@1 (FB) values reported in the main paper.}
\centering
\small
\setlength{\tabcolsep}{4pt}
\resizebox{\linewidth}{!}{%
\begin{tabular}{l|cc|cc|cc}
\toprule
& \multicolumn{2}{c|}{\textbf{Qiskit}}
& \multicolumn{2}{c|}{\textbf{Cirq}}
& \multicolumn{2}{c}{\textbf{PennyLane}} \\
\textbf{Model}
& \textbf{Pass@1} & \textbf{Pass@1 (FB)}
& \textbf{Pass@1} & \textbf{Pass@1 (FB)}
& \textbf{Pass@1} & \textbf{Pass@1 (FB)} \\
\midrule
\texttt{Gemini-3-Pro} & \textbf{59.5} & 73.8 & \textbf{54.8} & \textbf{76.2} & 40.5 & \textbf{66.7} \\
\texttt{GPT-5.1} & 57.1 & \textbf{83.3} & 52.4 & 73.8 & \textbf{42.9} & \textbf{66.7} \\
\texttt{DeepSeek-R1} & 50.0 & 71.4 & 45.2 & 61.9 & 33.3 & 52.4 \\
\texttt{MoonshotAI-Kimi-K2-Thinking} & 47.6 & 69.0 & 38.1 & 57.1 & 26.2 & 38.1 \\
\texttt{Claude-3.7-Sonnet} & 45.2 & 57.1 & 35.7 & 59.5 & 26.2 & 47.6 \\
\texttt{GPT-4.1} & 50.0 & 57.1 & 33.3 & 57.1 & 23.8 & 45.2 \\
\texttt{DeepSeek-Chat} & 45.2 & 42.9 & 28.6 & 40.5 & 31.0 & 45.2 \\
\texttt{Z-ai-GLM-4.7} & 42.9 & 69.0 & 38.1 & 61.9 & 23.8 & 64.3 \\
\texttt{Gemini-2.5-Flash} & 40.5 & 61.9 & 35.7 & 50.0 & 23.8 & 40.5 \\
\texttt{Llama-4-Maverick} & 38.1 & 54.8 & 28.6 & 42.9 & 19.0 & 38.1 \\
\texttt{MiniMax-M2.1} & 28.6 & 57.1 & 23.8 & 47.6 & 31.0 & 47.6 \\
\texttt{Qwen-2.5-7B-Instruct} & 16.7 & 19.0 & 4.8 & 7.1 & 11.9 & 19.0 \\
\bottomrule
\end{tabular}
}
\normalsize
\label{tab:pass1_by_framework}
\end{table}
\FloatBarrier

\section{Calibration of the KL Acceptance Threshold}
\label{sec:kl_threshold_proof}

Some benchmark tasks are probabilistic: correctness is defined by matching a target measurement distribution rather than a single deterministic output. Even the canonical circuit exhibits finite-shot variability, so we calibrate the acceptance threshold from repeated canonical executions instead of setting it heuristically.

For each probabilistic task $t$, we run the canonical reference circuit $R=1000$ times to obtain empirical distributions $\{P^{(t)}_i\}_{i=1}^{R}$ and define the task reference distribution as their renormalized mean:
\begin{equation}
P^{(t)}_{\mathrm{ref}} = \mathrm{Normalize}\!\left(\frac{1}{R}\sum_{i=1}^{R} P^{(t)}_i\right).
\end{equation}

We then measure within-canonical variability via the null KL distribution
\begin{equation}
d^{(t)}_i = D_{\mathrm{KL}}\!\left(\widetilde{P}^{(t)}_{\mathrm{ref}} \,\middle\|\, \widetilde{P}^{(t)}_i\right),
\end{equation}
and select a global threshold from a high quantile of the pooled null values across tasks:
\begin{equation}
\tau_{\mathrm{global}}(q)=\mathrm{Quantile}_{q}\!\left(\bigcup_t \{d^{(t)}_i\}_{i=1}^{R}\right).
\end{equation}

With $q=0.997$, the resulting pooled threshold is $\tau_{\mathrm{global}}=0.048$, so we use $\tau=0.05$ as a slightly more permissive paper-wide constant.

\begin{figure}[htbp]
\centering
\includegraphics[width=0.64\linewidth]{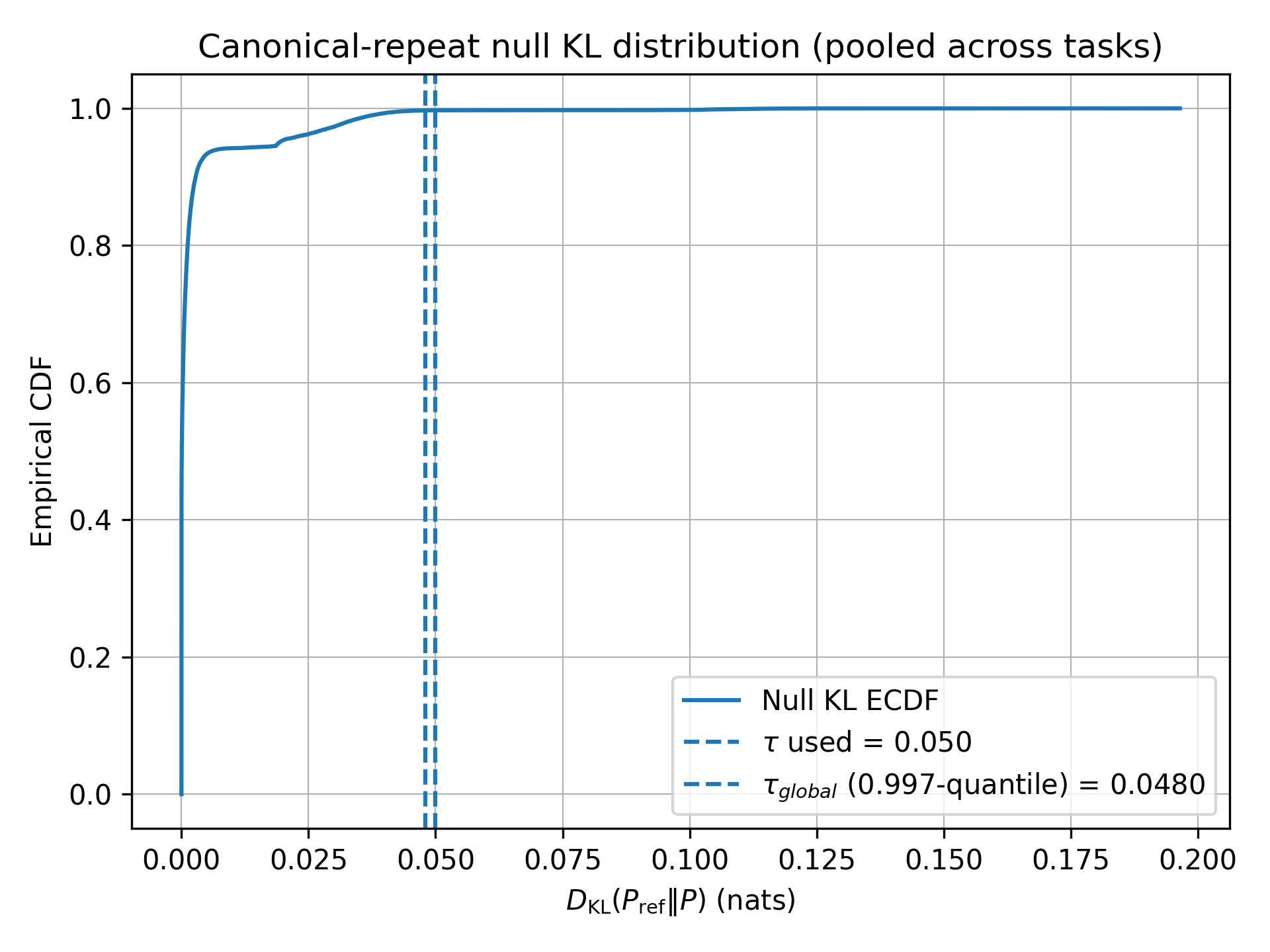}
\caption{\textbf{The null KL distribution supports the global acceptance threshold.} The pooled canonical-repeat ECDF places the 99.7th percentile at 0.048, motivating the paper-wide threshold $\tau=0.05$.}
\label{fig:null_kl_ecdf}
\end{figure}
\FloatBarrier

\section{Task Categories and Examples}
\label{sec:app_task_categories}

\begin{table}[H]
\caption{\textbf{Quantum algorithms dominate the benchmark mix.} \qbplus contains 42 tasks, with most concentrated in algorithmic reasoning.}
\centering
\begin{tabular}{lc}
\toprule
\textbf{Category} & \textbf{Number of Tasks}\\
\midrule
Quantum Algorithms & 31\\
State Preparation & 6\\
Decomposition & 5\\
\midrule
\textbf{Total} & \textbf{42}\\
\bottomrule
\end{tabular}
\label{tab:tasks_categories}
\end{table}

\qbplus organizes tasks equivalently across frameworks:

\begin{itemize}
    \item \textbf{Quantum Algorithms}: implement known algorithms or subroutines.
    \item \textbf{Gate Decomposition}: convert high-level operations into native gates.
    \item \textbf{State Preparation}: construct circuits to produce target quantum states.
\end{itemize}

\section{Pass@1 vs Pass@5 Comparisons}
\label{sec:supp_pass1_pass5}

Pass@1 measures top-1 solution correctness, while Pass@5 measures correctness across the top 5 generated solutions.

\noindent\textbf{What to look for:} These figures show whether correct solutions are absent altogether or simply not selected on the first try. Large gaps between Pass@1 and Pass@5 indicate that models often contain the right solution among a small set of samples, even when one-shot decoding misses it.
\FloatBarrier
\begin{figure}[htbp]
\centering
\includegraphics[width=\linewidth]{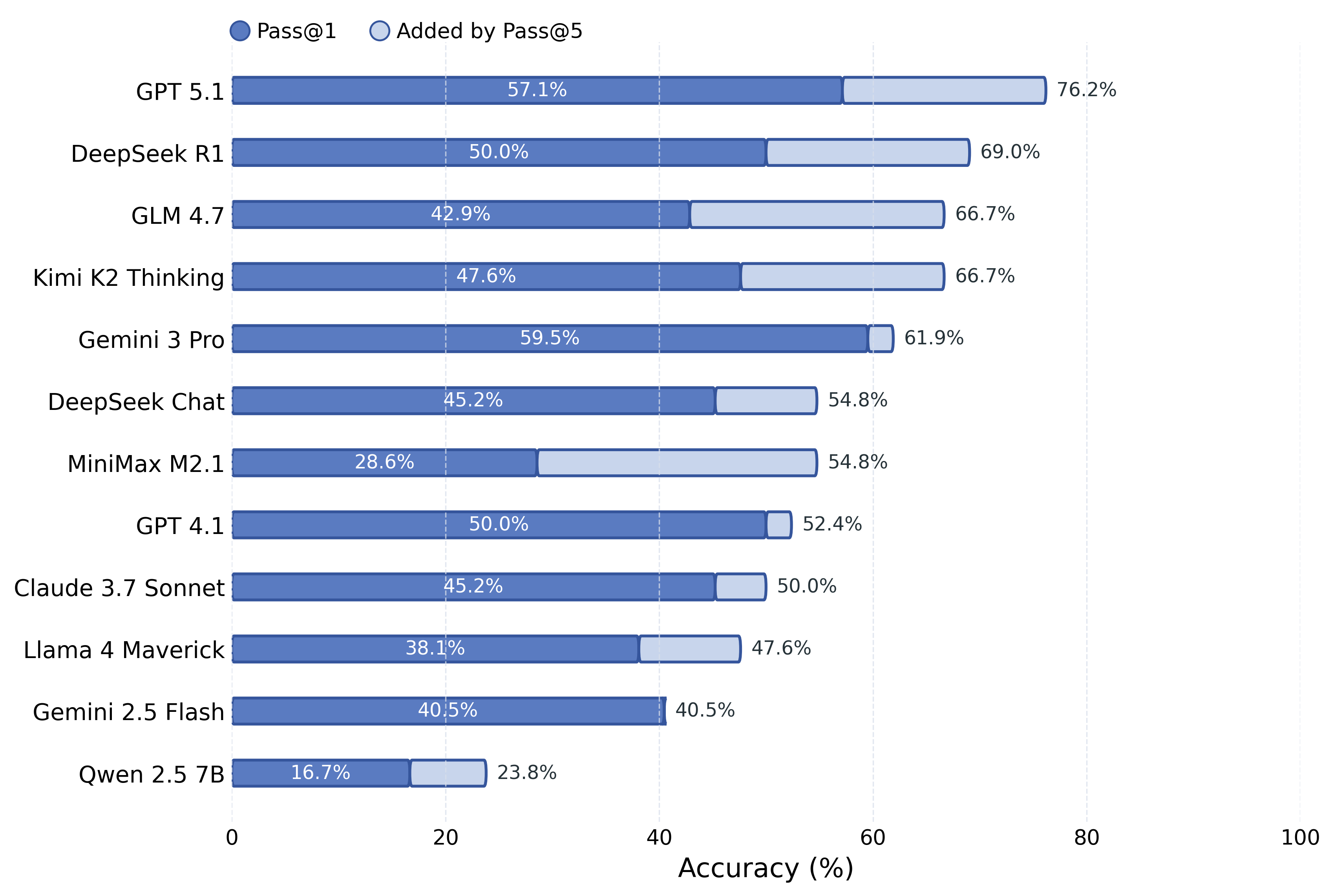}
\caption{\textbf{Multiple samples recover additional Qiskit solutions.} The gap between Pass@1 and Pass@5 identifies tasks where one-shot decoding leaves recoverable performance on the table.}
\label{fig:pass1_vs_pass5_qiskit}
\end{figure}
\FloatBarrier
\begin{figure}[htbp]
\centering
\includegraphics[width=\linewidth]{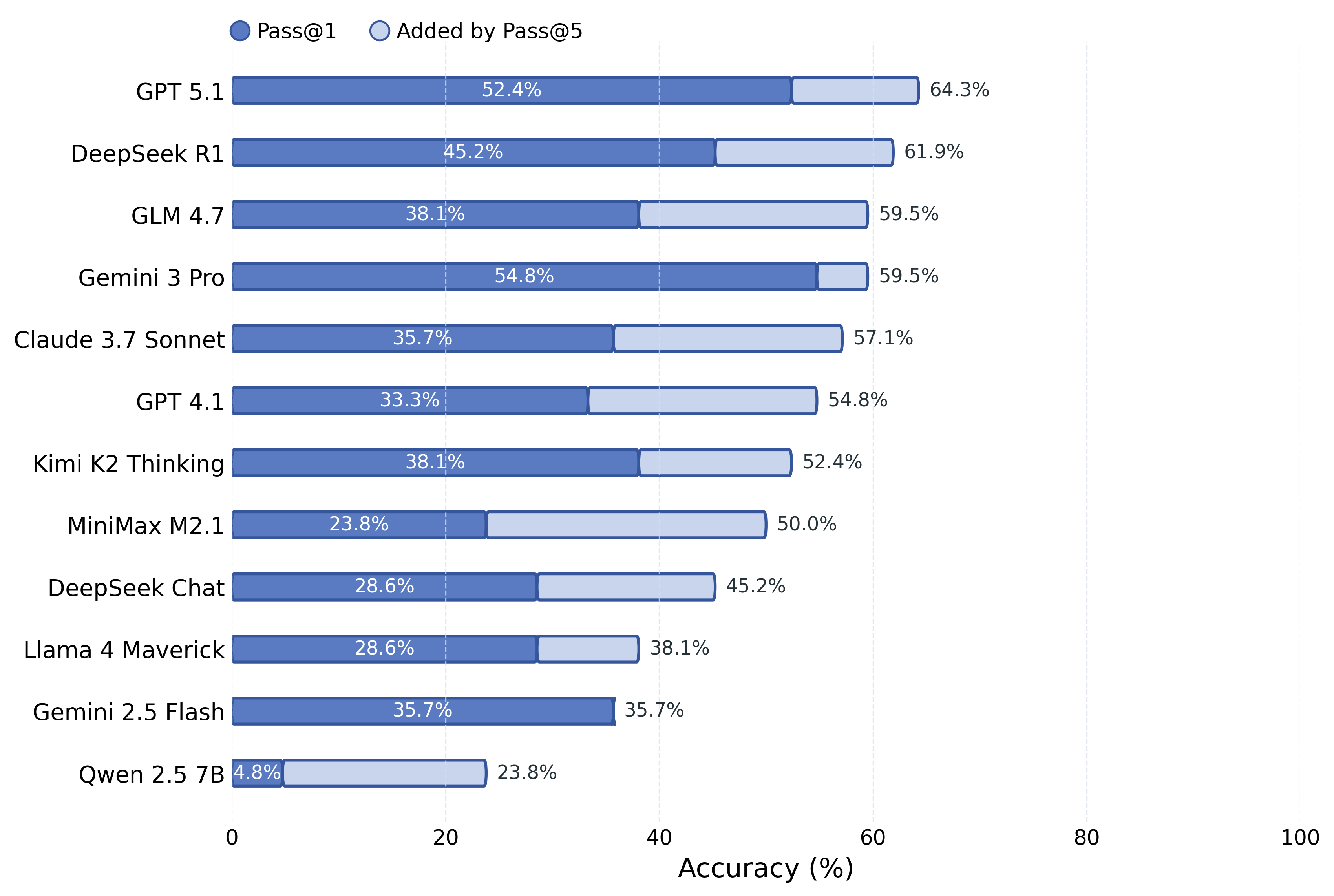}
\caption{\textbf{Cirq also benefits meaningfully from multi-sample generation.} The gains are especially visible among the middle of the model ranking.}
\label{fig:pass1_vs_pass5_cirq}
\end{figure}

\begin{figure}[htbp]
\centering
\includegraphics[width=\linewidth]{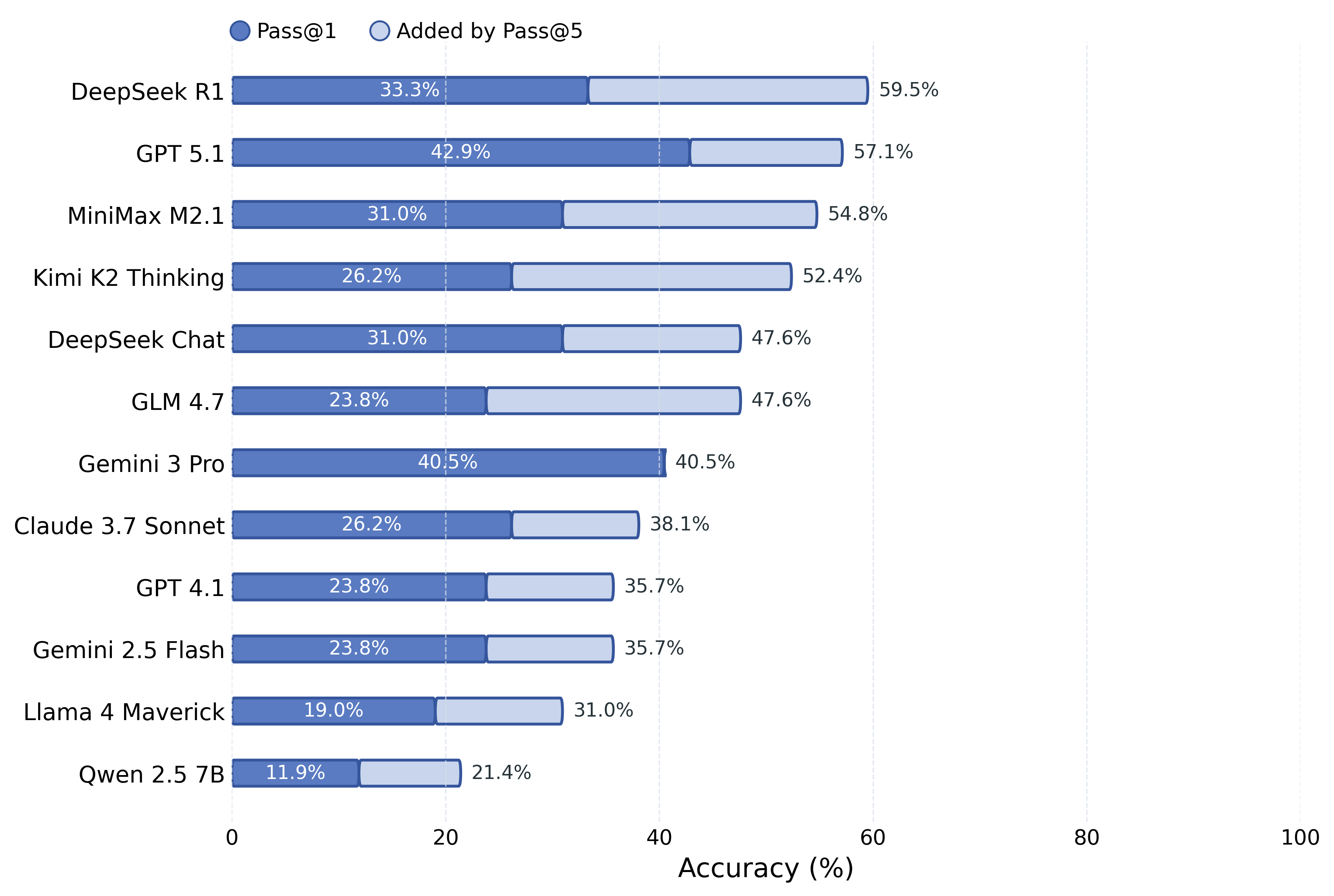}
\caption{\textbf{PennyLane retains large recoverable gaps for weaker models.} Multi-sample decoding helps, but it does not close the framework-level difficulty gap.}
\label{fig:pass1_vs_pass5_Pennylane}
\end{figure}

\FloatBarrier

\section{Per-Task Heatmaps}
\label{sec:supp_heatmaps}

\subsection*{Pass@1 Heatmaps}

\noindent\textbf{What to look for:} The Pass@1 heatmaps show where one-shot reliability is genuinely strong and where it breaks down task by task. Dense horizontal bands indicate broadly capable models; persistent white columns indicate tasks that remain difficult for almost everyone.

\FloatBarrier
\begin{figure}[htbp]
\centering
\includegraphics[width=\linewidth]{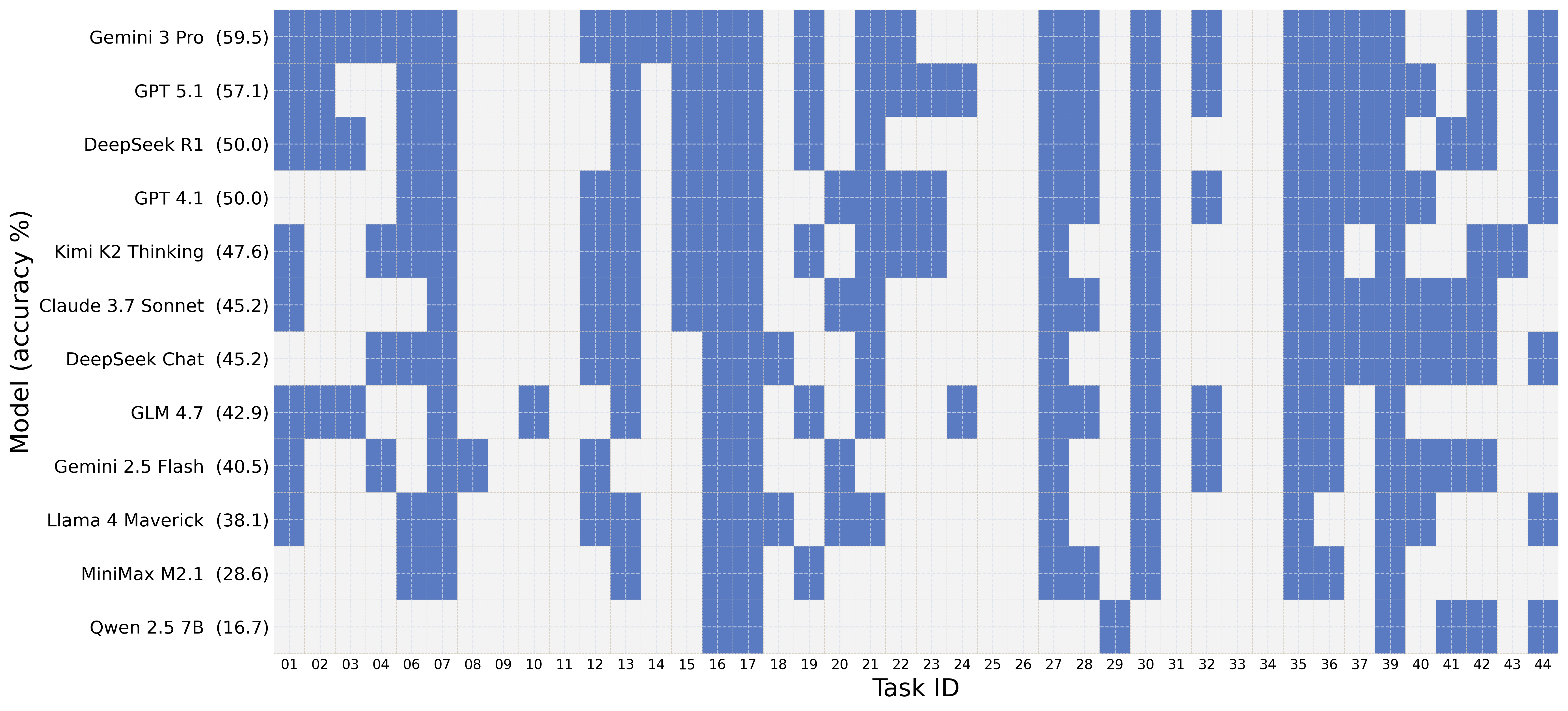}
\caption{\textbf{One-shot success in Qiskit is concentrated in a broad but incomplete task band.} Each row corresponds to a model and each column to a task.}
\label{fig:app_pass1_qiskit}
\end{figure}
\FloatBarrier
\begin{figure}[htbp]
\centering
\includegraphics[width=\linewidth]{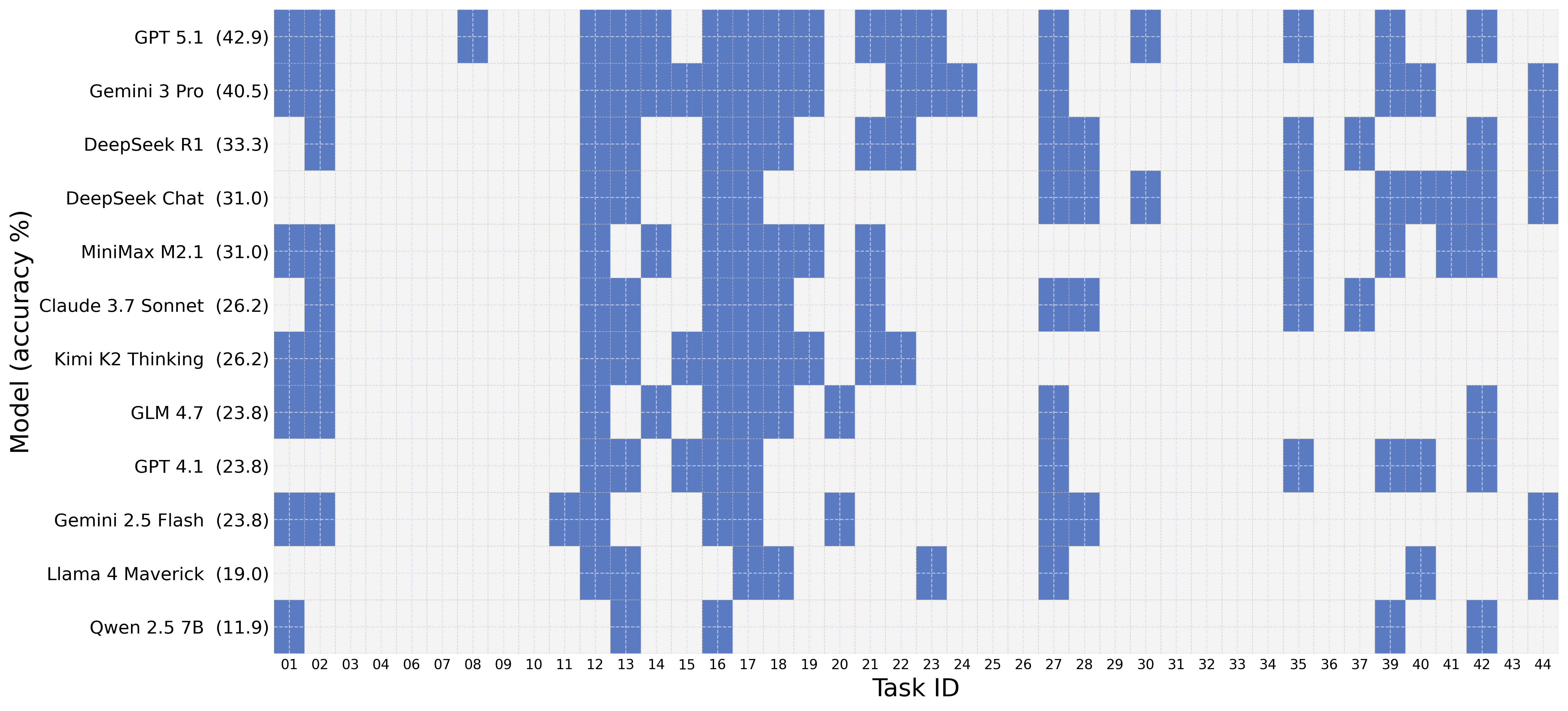}
\caption{\textbf{PennyLane exposes a noticeably sparser one-shot success map.} Each row corresponds to a model and each column to a task.}
\label{fig:app_pass1_Pennylane}
\end{figure}
\FloatBarrier
\begin{figure}[htbp]
\centering
\includegraphics[width=\linewidth]{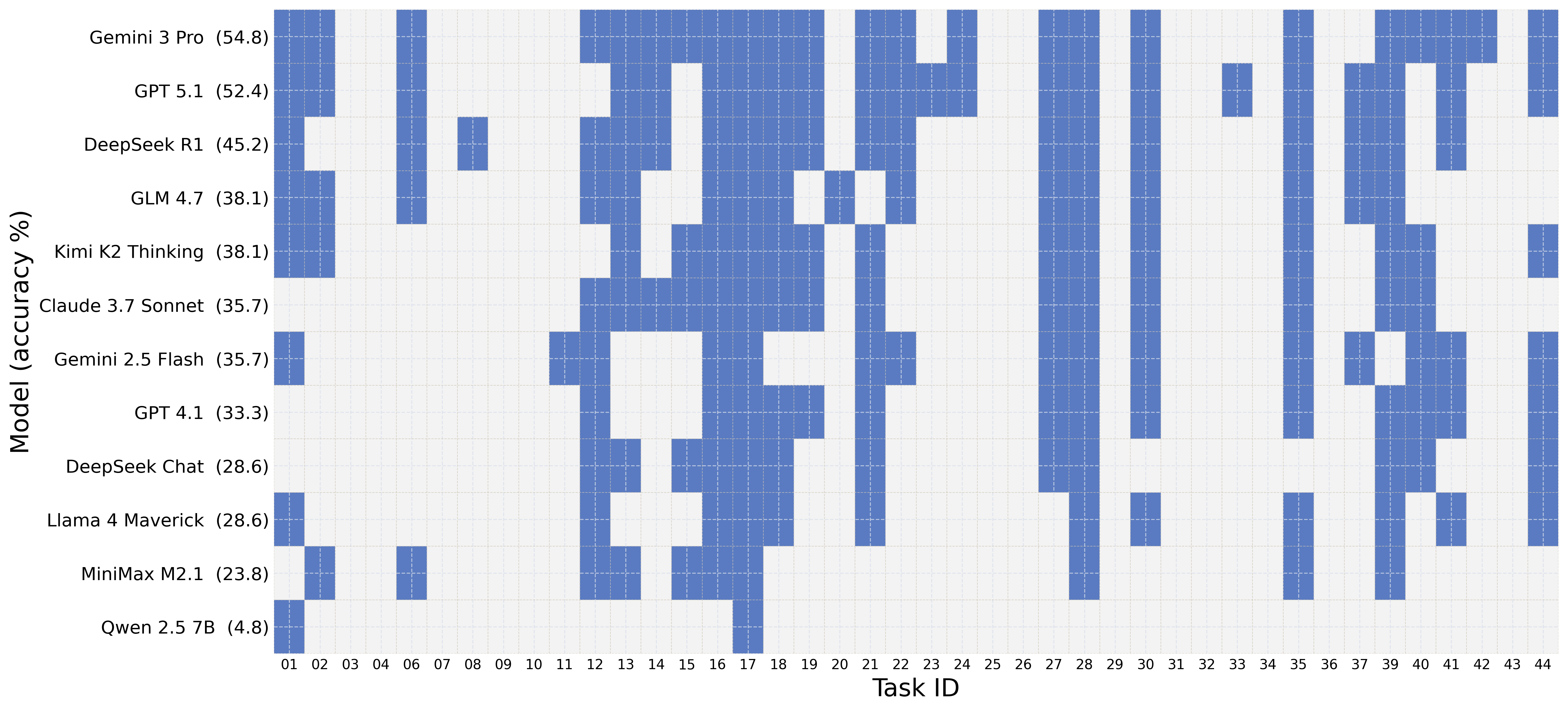}
\caption{\textbf{Cirq sits between Qiskit and PennyLane in first-attempt density.} The overall pattern is stronger than PennyLane but less complete than Qiskit.}
\label{fig:app_pass1_cirq_heatmap}
\end{figure}
\FloatBarrier
\subsection*{Pass@5 Heatmaps}

\noindent\textbf{What to look for:} Compared with the Pass@1 maps, these heatmaps reveal how much additional coverage appears once models are allowed multiple tries. New dark regions indicate tasks where the capability exists but is unstable under one-shot decoding.

\begin{figure}[htbp]
\centering
\includegraphics[width=\linewidth]{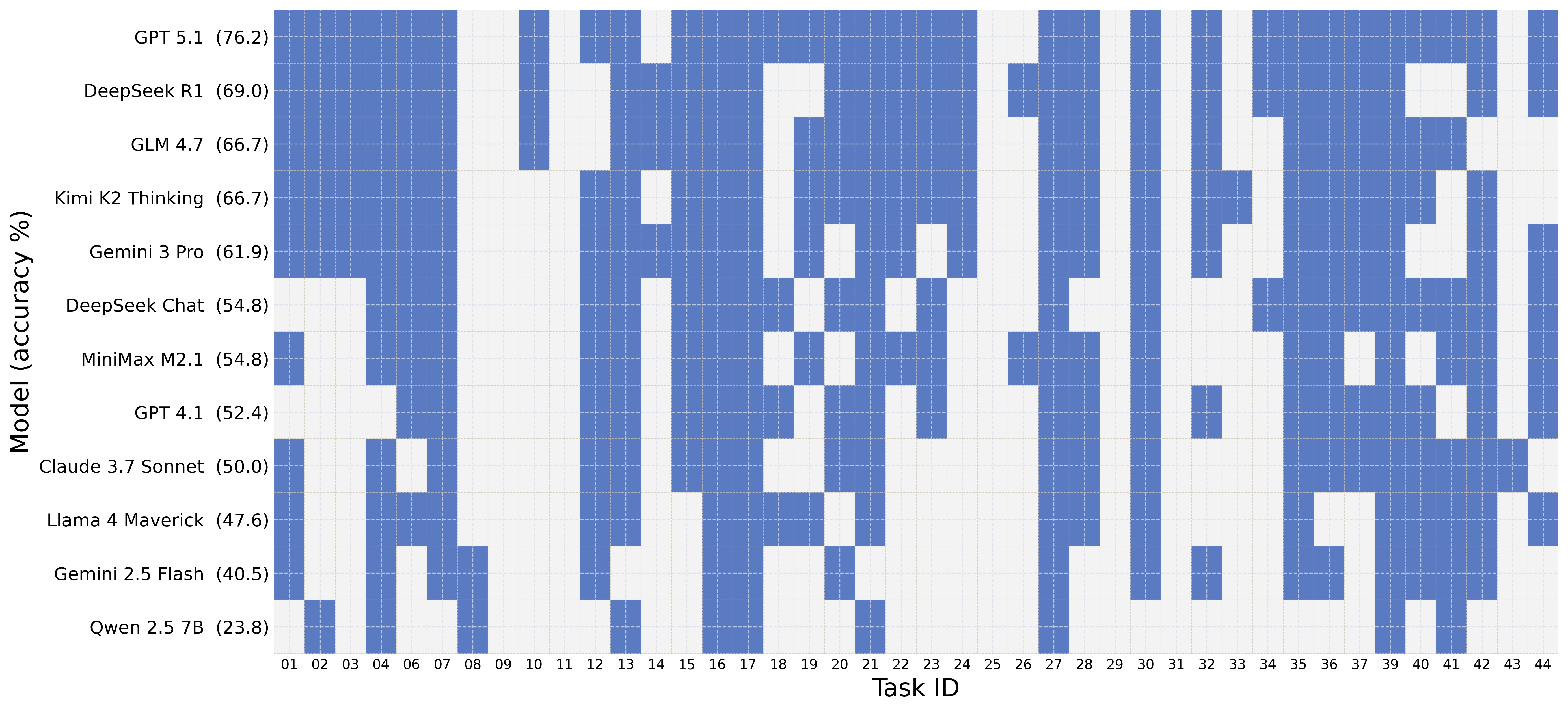}
\caption{\textbf{Pass@5 broadens Qiskit coverage substantially.} Multi-sample decoding turns many partial one-shot failures into recoverable successes.}
\label{fig:app_pass5_qiskit}
\end{figure}
\FloatBarrier
\begin{figure}[htbp]
\centering
\includegraphics[width=\linewidth]{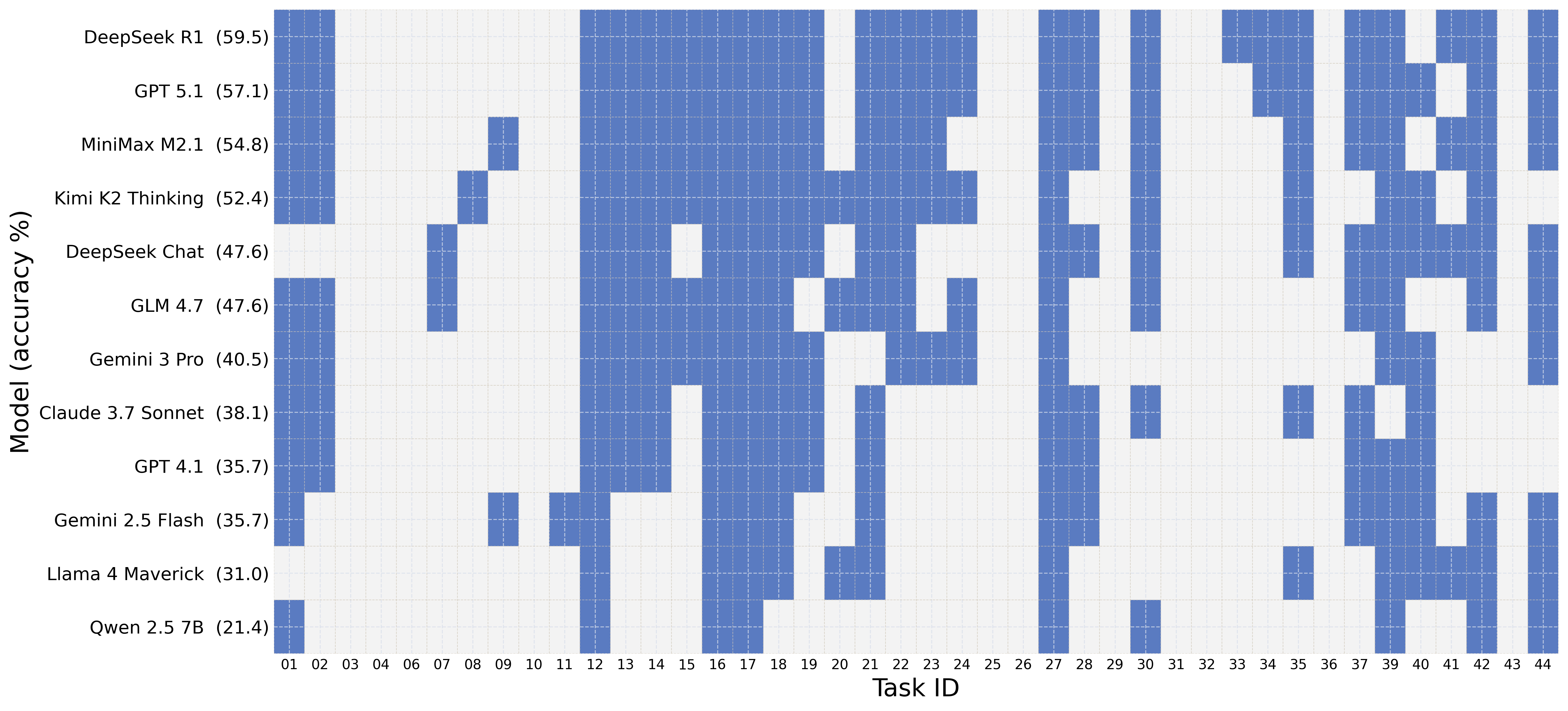}
\caption{\textbf{Pass@5 helps in PennyLane, but hard tasks remain visibly persistent.} Multi-sample decoding broadens coverage without removing the framework gap.}
\label{fig:app_pass5_Pennylane}
\end{figure}
\FloatBarrier
\begin{figure}[htbp]
\centering
\includegraphics[width=\linewidth]{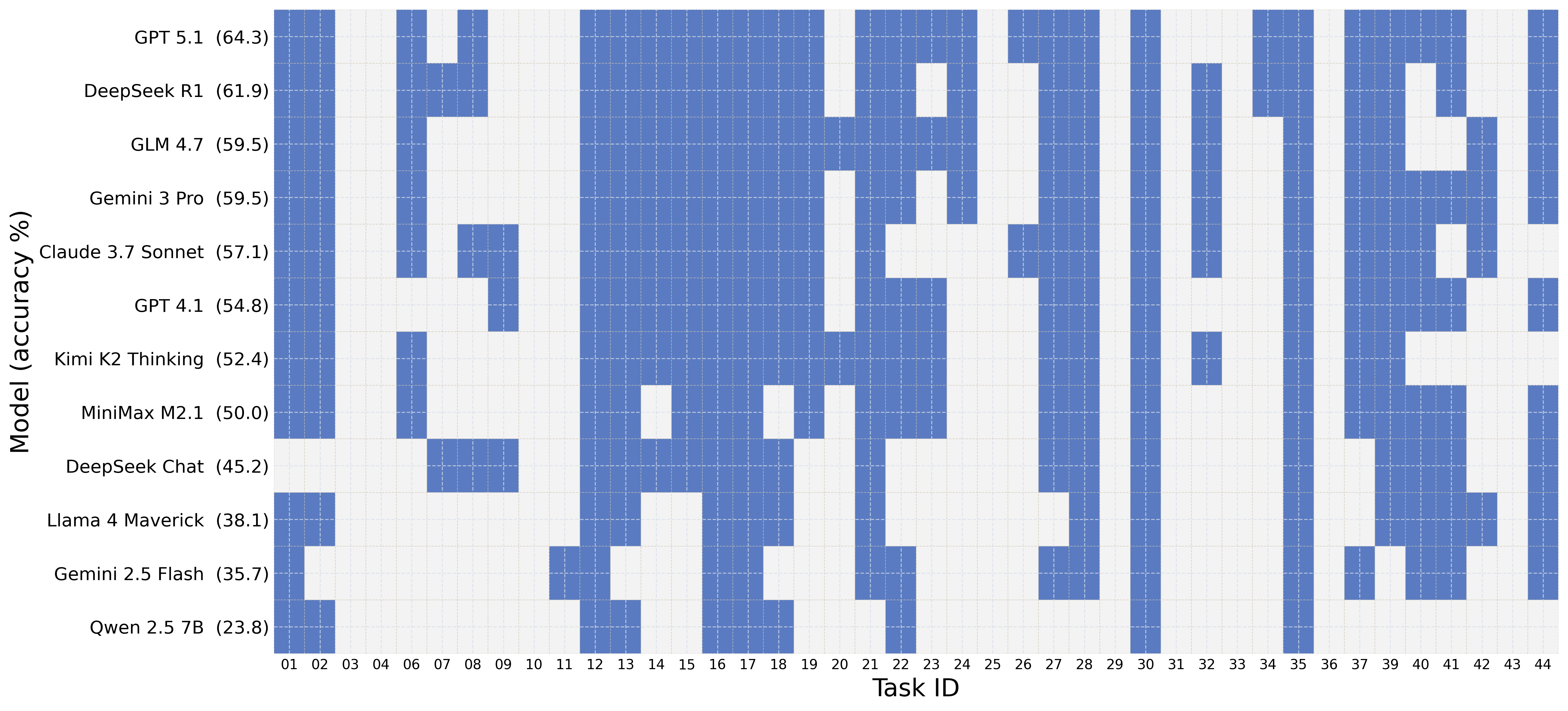}
\caption{\textbf{Cirq gains a wider solvable region under Pass@5.} The additional coverage confirms that many one-shot failures are unstable rather than absolute.}
\label{fig:app_pass5_cirq}
\end{figure}
\FloatBarrier
\begin{table}[H]
\caption{\textbf{Pass@5 narrows but does not remove framework gaps.} Accuracy (\%) over benchmark tasks for each framework.}
\centering
\begin{tabular}{lccc}
\toprule
\textbf{Model} & \textbf{Qiskit} & \textbf{Cirq} & \textbf{PennyLane} \\
\midrule
\texttt{Gemini-3-Pro} & 61.9 & 59.5 & 40.5 \\
\texttt{GPT-5.1} & \textbf{76.2} & \textbf{64.3} & 57.1 \\
\texttt{DeepSeek-R1} & 69.0 & 61.9 & \textbf{59.5} \\
\texttt{MoonshotAI-Kimi-K2-Thinking} & 66.7 & 52.4 & 52.4 \\
\texttt{Claude-3.7-Sonnet} & 50.0 & 57.1 & 38.1 \\
\texttt{GPT-4.1} & 52.4 & 54.8 & 35.7 \\
\texttt{DeepSeek-Chat} & 54.8 & 45.2 & 47.6 \\
\texttt{Z-ai-GLM-4.7} & 66.7 & 59.5 & 47.6 \\
\texttt{Gemini-2.5-Flash} & 40.5 & 35.7 & 35.7 \\
\texttt{Llama-4-Maverick} & 47.6 & 38.1 & 31.0 \\
\texttt{MiniMax-M2.1} & 54.8 & 50.0 & 54.8 \\
\texttt{Qwen-2.5-7B-Instruct} & 23.8 & 23.8 & 21.4 \\
\bottomrule
\end{tabular}
\label{tab:app_pass5_by_framework}
\end{table}
\FloatBarrier
\section{Prefill vs No-Prefill}
\label{sec:supp_prefill_noprefill}

We evaluate two prompting conditions for all models and frameworks:

\begin{itemize}
    \item \textbf{Prefill}: the prompt includes required imports, function signature, and minimal boilerplate.
    \item \textbf{No-prefill}: the model generates the full solution from scratch.
\end{itemize}

\noindent\textbf{What to look for:} These figures isolate how much of the error budget comes from boilerplate and setup rather than task logic. Larger gaps between the paired bars indicate models that depend heavily on scaffolding to produce executable framework code.

\begin{figure}[htbp]
\centering
\includegraphics[width=\linewidth]{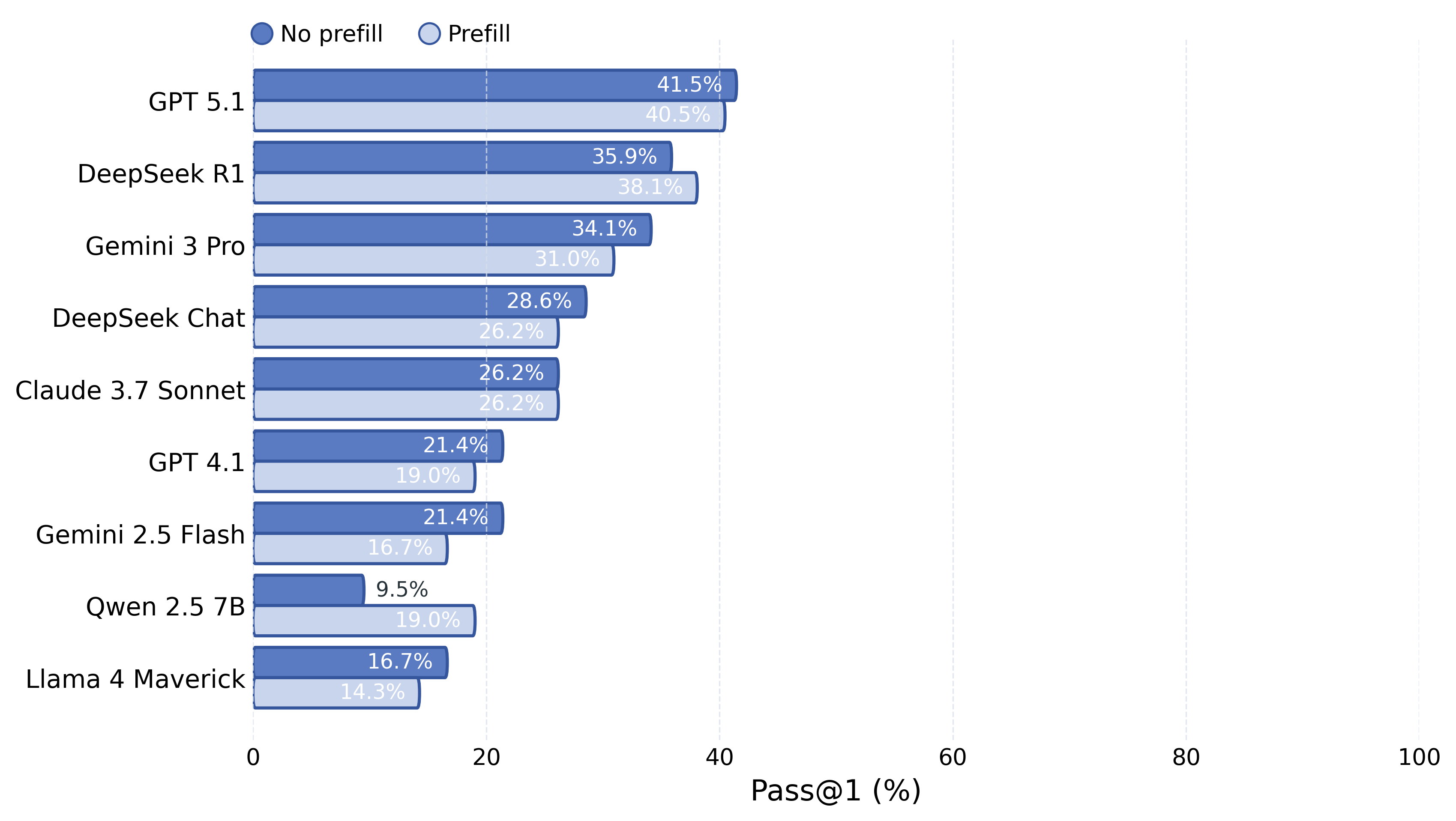}
\caption{\textbf{Prefill helps most when PennyLane boilerplate is easy to miss.} The ranking changes confirm that setup friction still matters for several mid-tier models.}
\label{fig:supp_prefill_Pennylane}
\end{figure}

\FloatBarrier
\begin{figure}[htbp]
\centering
\includegraphics[width=\linewidth]{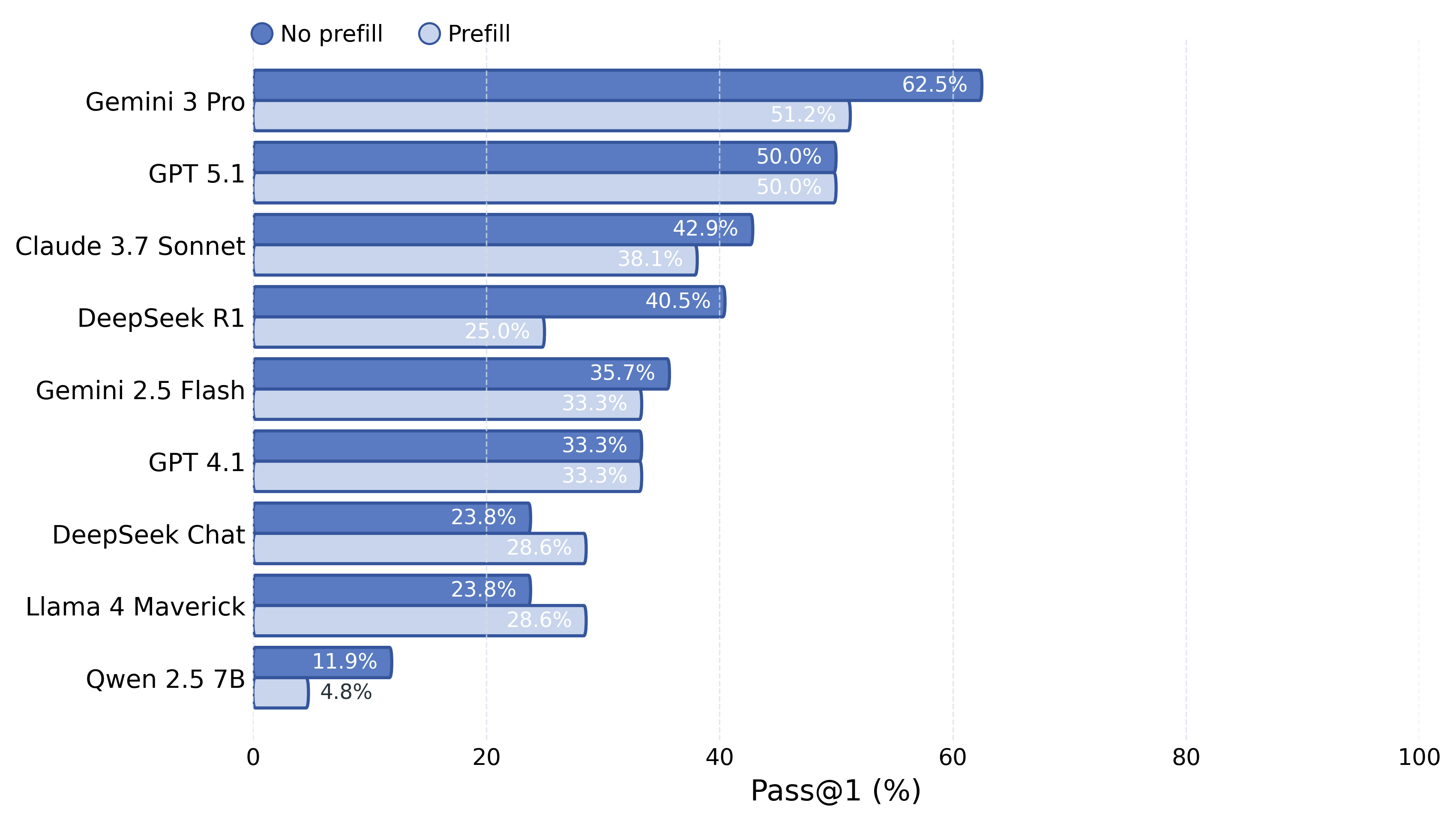}
\caption{\textbf{Cirq also shows meaningful sensitivity to prompt scaffolding.} Prefill changes both average accuracy and several mid-tier rankings.}
\label{fig:supp_prefill_cirq}
\end{figure}
\FloatBarrier
\begin{figure}[htbp]
\centering
\includegraphics[width=\linewidth]{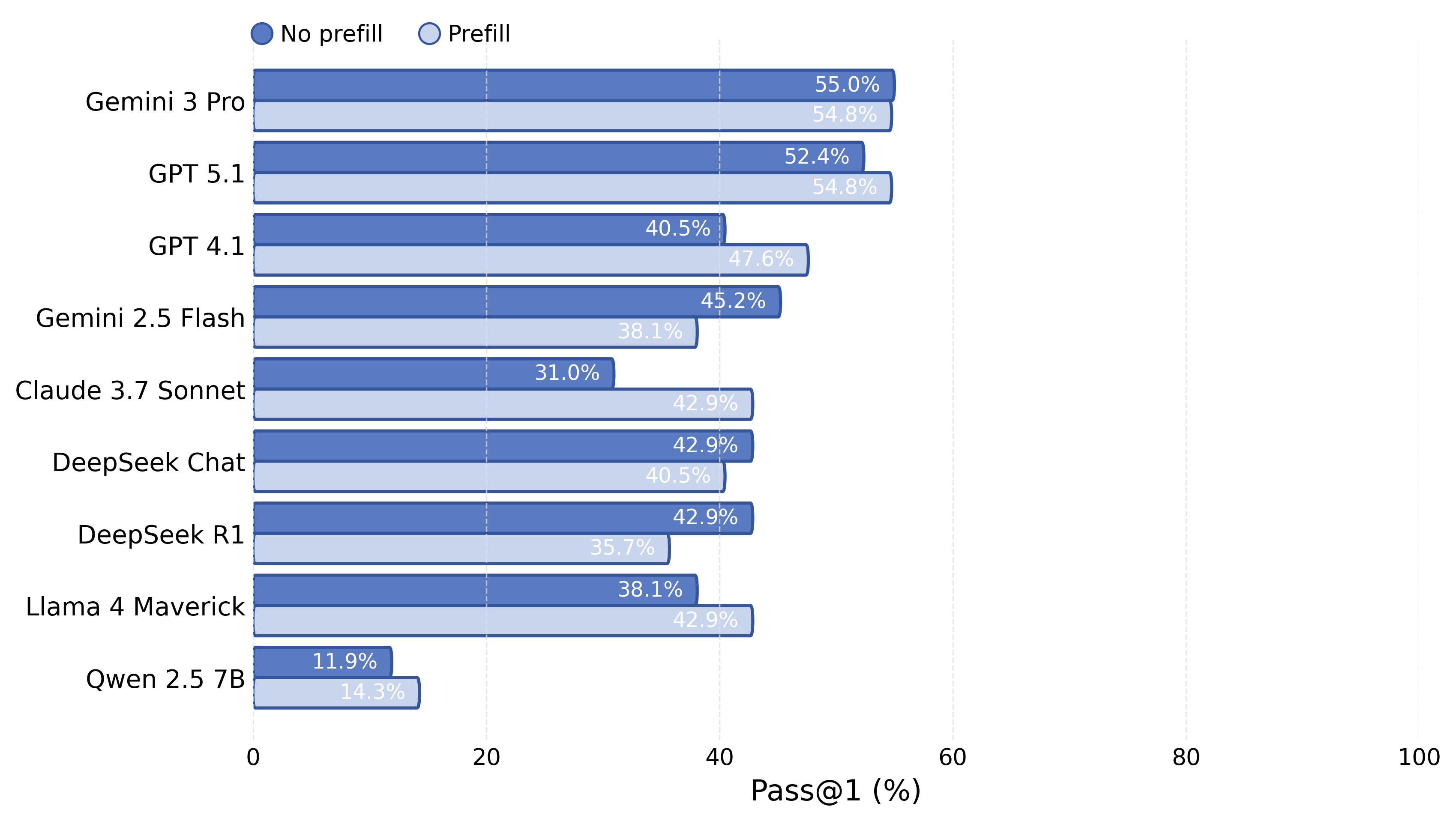}
\caption{\textbf{Qiskit benefits from prefill, but less uniformly than weaker frameworks.} The effect is real, though not consistent across the full model range.}
\label{fig:supp_prefill_qiskit}
\end{figure}

\FloatBarrier

\section{Error Distributions}
\label{sec:supp_error}
This section examines what goes wrong when first-attempt solutions fail. The goal is to separate semantic mistakes from implementation and framework-use errors.
\begin{figure}[htbp]
\centering
\includegraphics[width=\linewidth]{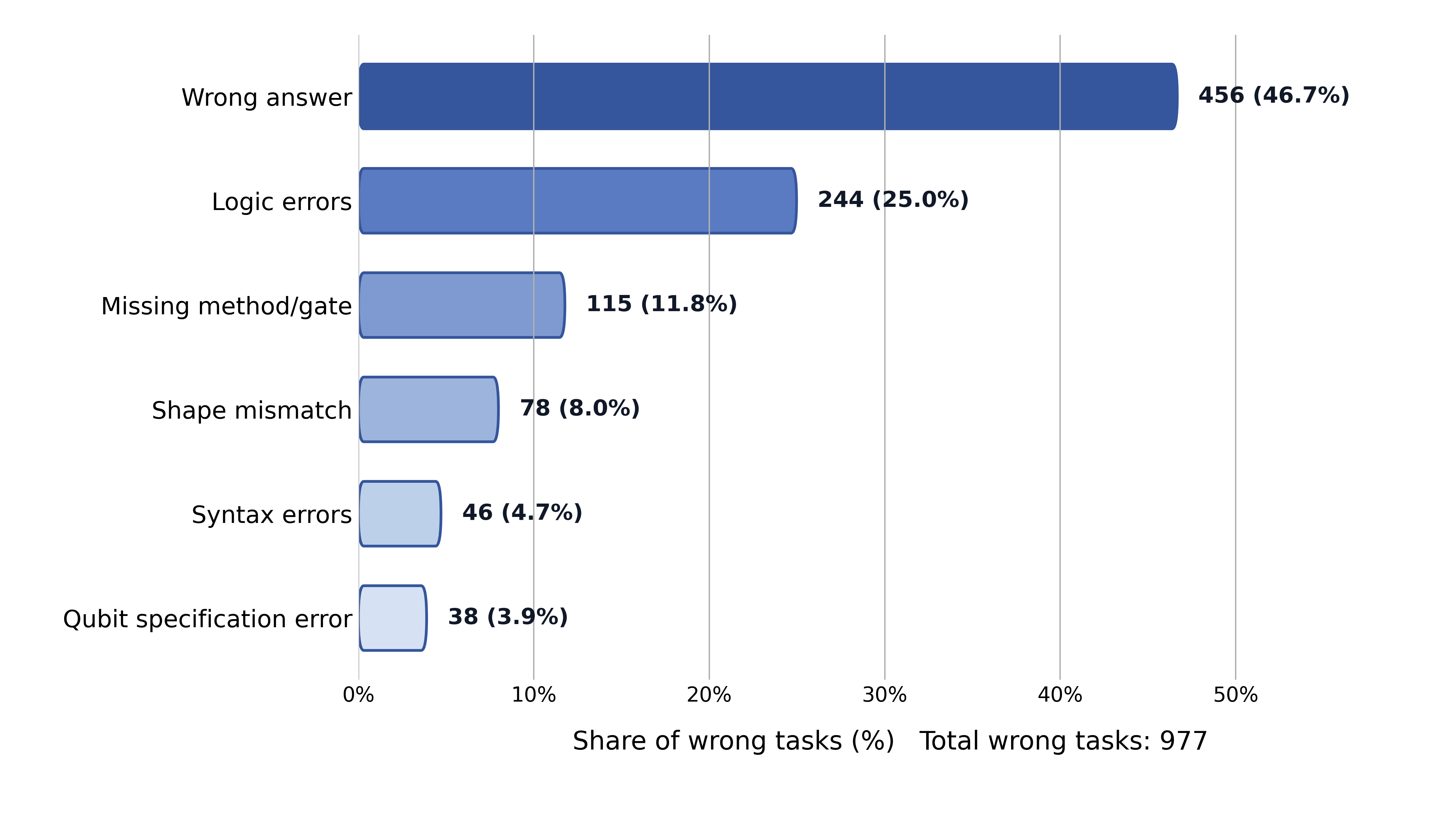}
\caption{\textbf{Most first-attempt failures are semantic, not syntactic.} Wrong answers and logic errors dominate the Pass@1 error budget across frameworks.}
\label{fig:app_pass1_error_pie}
\end{figure}
\noindent\textbf{Observation:}
Figure~\ref{fig:app_pass1_error_pie} shows that most Pass@1 failures are driven by semantic mistakes: wrong answers (\(46.7\%\)) and logic errors (\(25.0\%\)) together dominate the error budget. More direct implementation problems still matter, but they are secondary, including missing methods/gates (\(11.8\%\)), shape mismatches (\(8.0\%\)), syntax errors (\(4.7\%\)), and qubit specification errors (\(3.9\%\)). This split helps explain why feedback can recover many first-attempt failures without eliminating the deeper reasoning gap.

\FloatBarrier

\section{Feedback-Loop Results}
\label{sec:supp_feedback}

We applied up to 5 repair attempts via feedback loops.

\noindent\textbf{What to look for:} The feedback plots show both the upside and the limit of iterative repair. Dense heatmaps and rapidly rising curves indicate that many failures are fixable once the model sees an execution signal, while the remaining gaps reveal the tasks that stay hard even after several retries.

\begin{figure}[htbp]
\centering
\includegraphics[width=\linewidth]{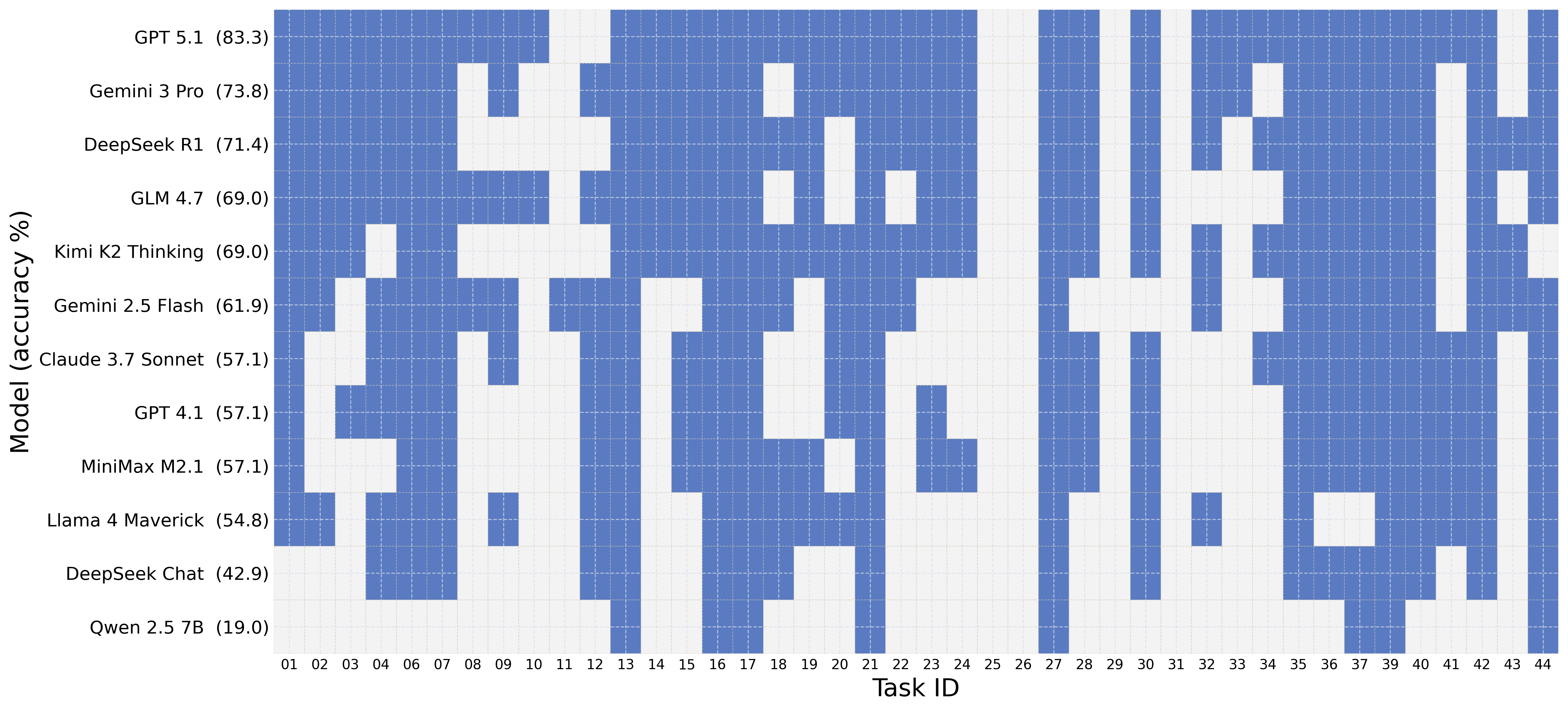}
\caption{\textbf{Feedback densifies the Qiskit success map.} Stronger models in particular convert many previously sparse regions into solved tasks.}
\label{fig:app_pass1fb_qiskit_heatmap}
\end{figure}

\begin{figure}[htbp]
\centering
\includegraphics[width=\linewidth]{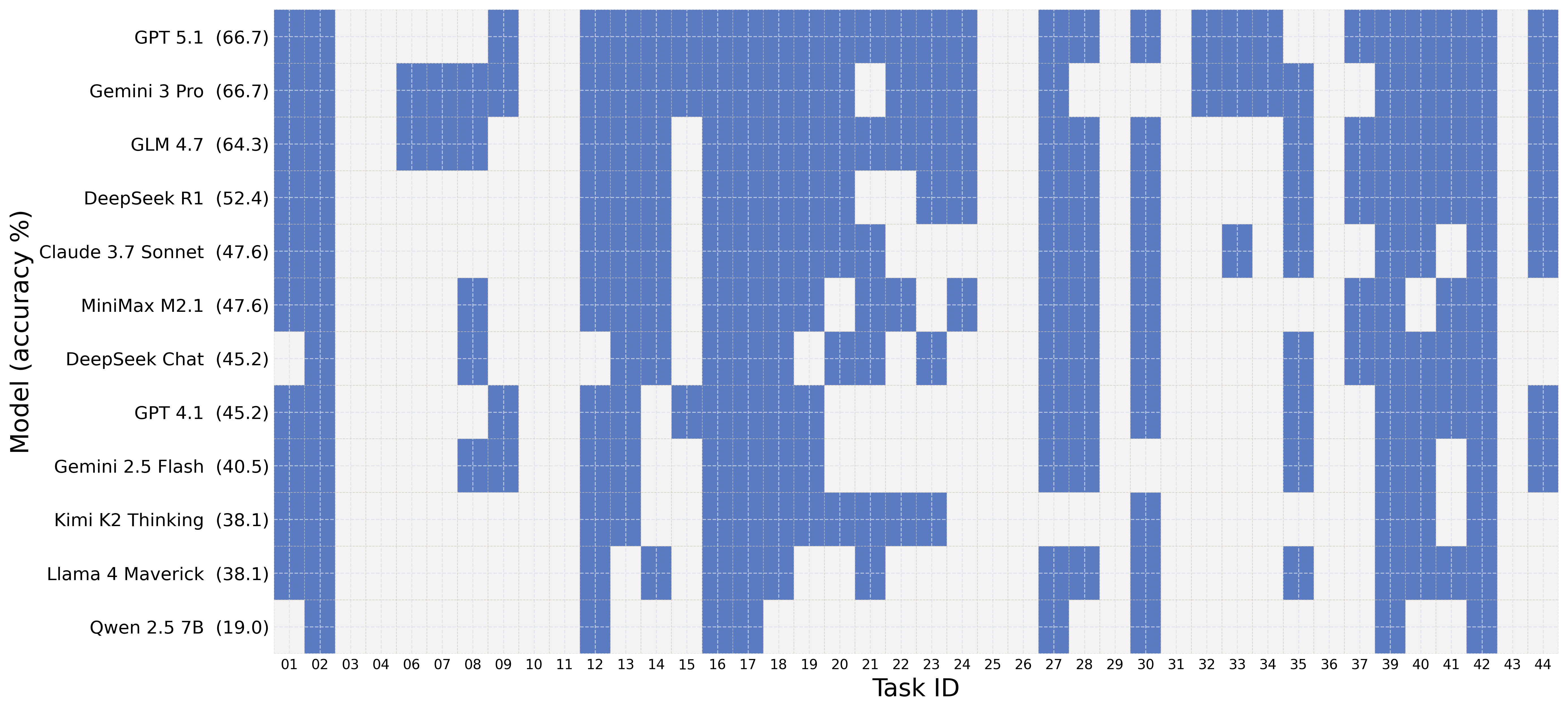}
\caption{\textbf{Feedback improves PennyLane coverage, but the map remains visibly harder.} The gains are substantial without fully closing the framework gap.}
\label{fig:app_pass1fb_Pennylane}
\end{figure}

\begin{figure}[htbp]
\centering
\includegraphics[width=\linewidth]{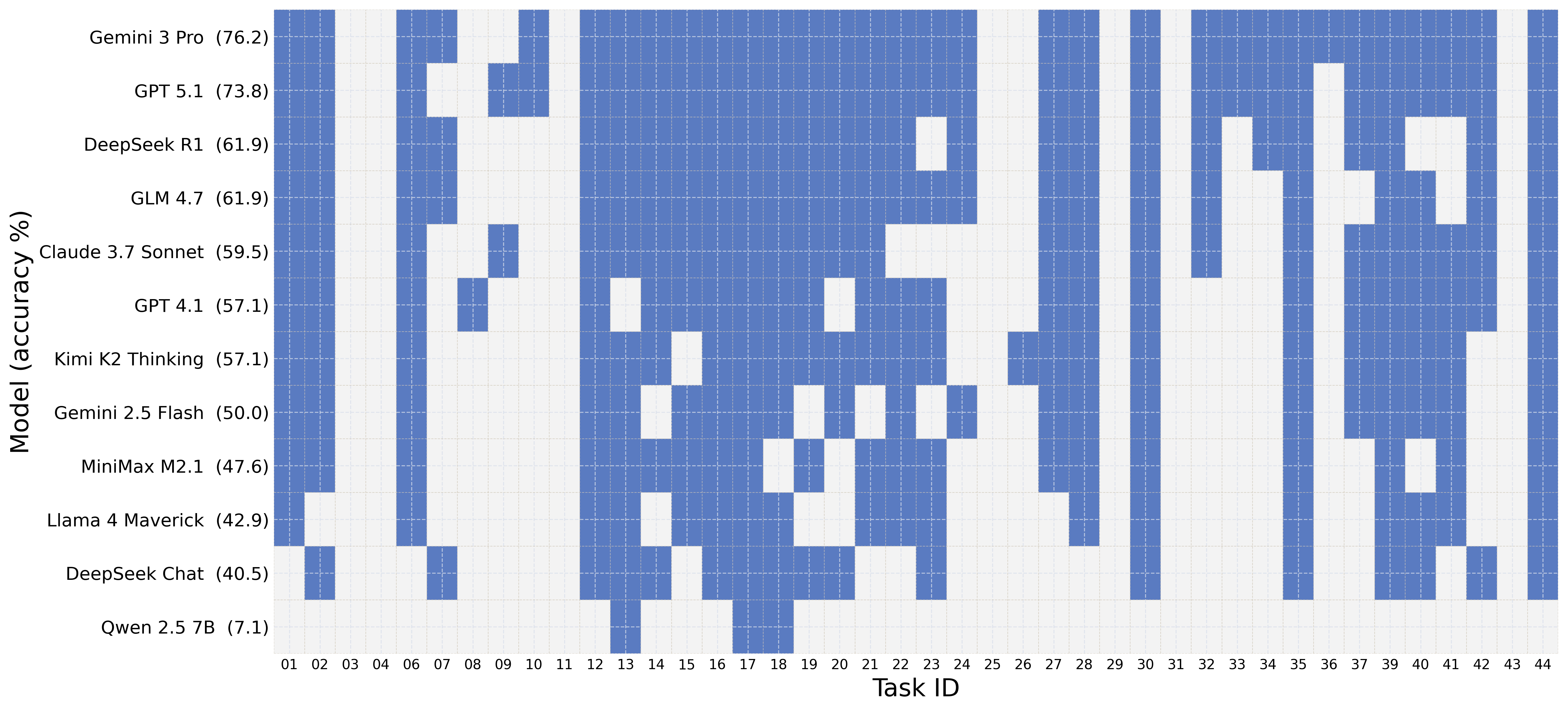}
\caption{\textbf{Feedback broadens Cirq success across much of the ranking.} The densification is clear, especially among stronger and mid-tier models.}
\label{fig:app_pass1fb_cirq__heatmap}
\end{figure}

\begin{figure}[htbp]
\centering
\includegraphics[width=\linewidth]{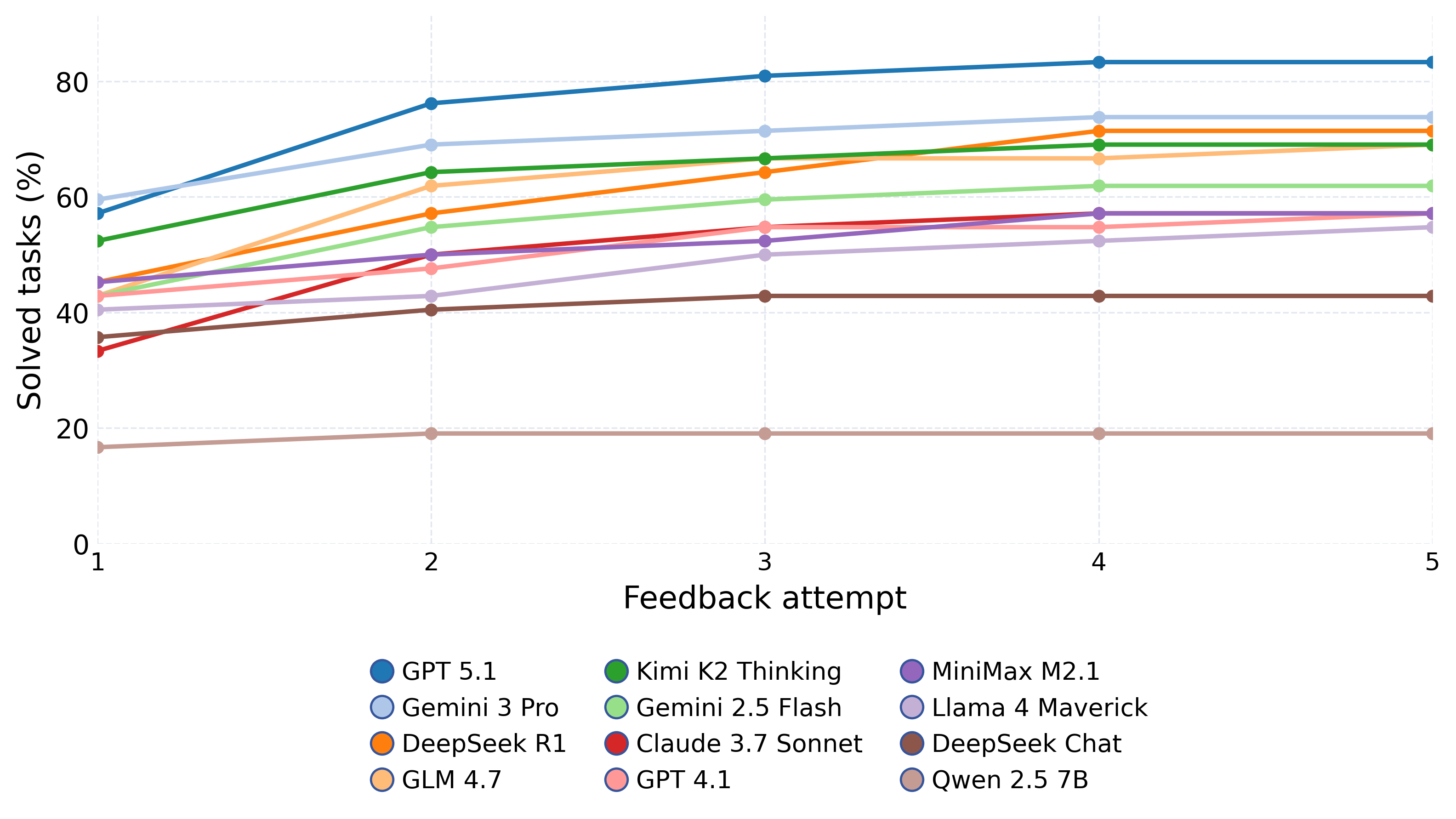}
\caption{\textbf{Most Qiskit feedback gains arrive early.} The curves rise quickly in the first repair rounds and then flatten.}
\label{fig:app_pass1fb_qiskit}
\end{figure}

\begin{figure}[htbp]
\centering
\includegraphics[width=\linewidth]{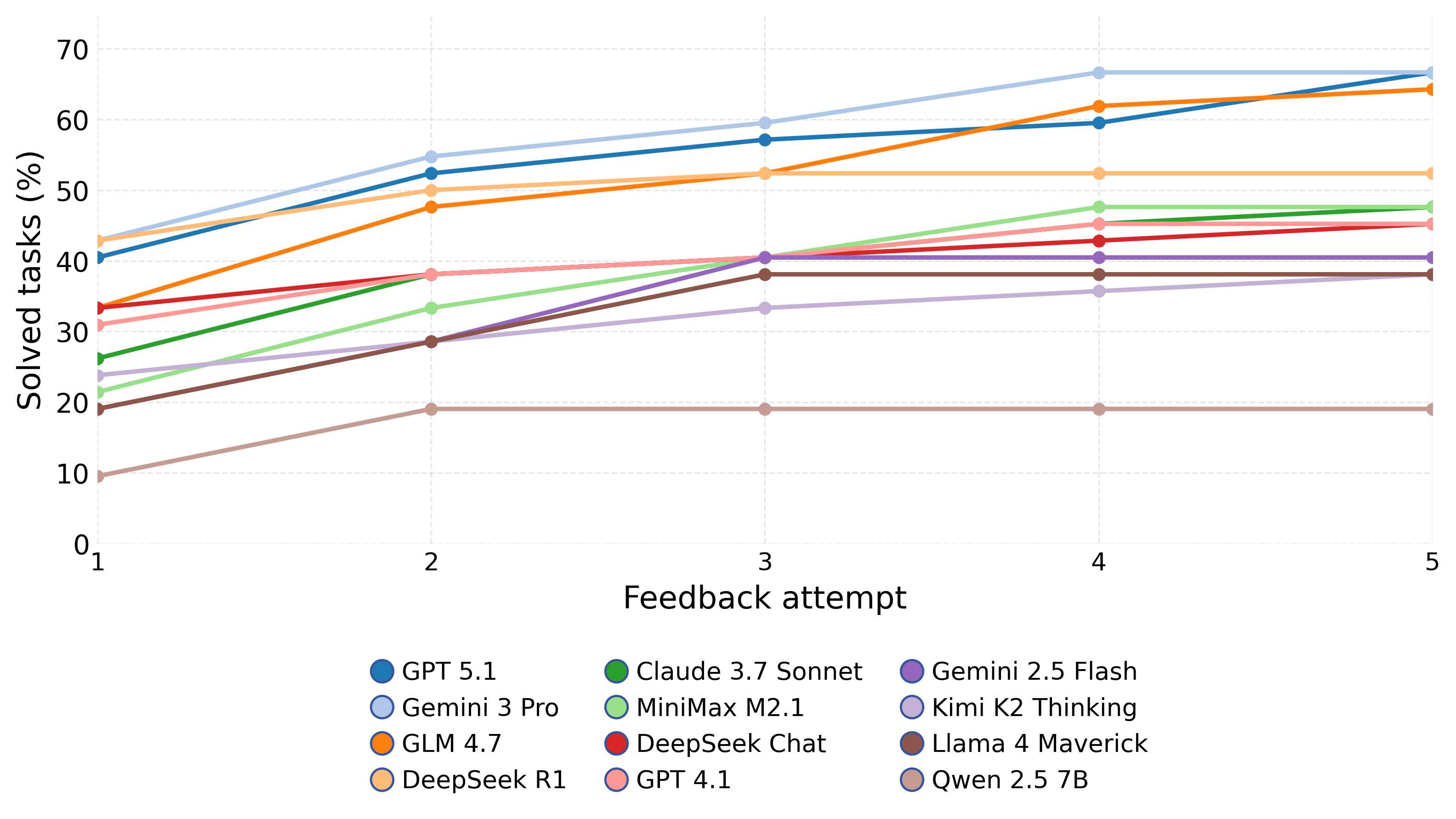}
\caption{\textbf{PennyLane improves steadily, but not indefinitely, with additional repair attempts.} Most of the lift still arrives in the early rounds.}
\label{fig:app_pass1fb_pennylane}
\end{figure}

\begin{figure}[htbp]
\centering
\includegraphics[width=\linewidth]{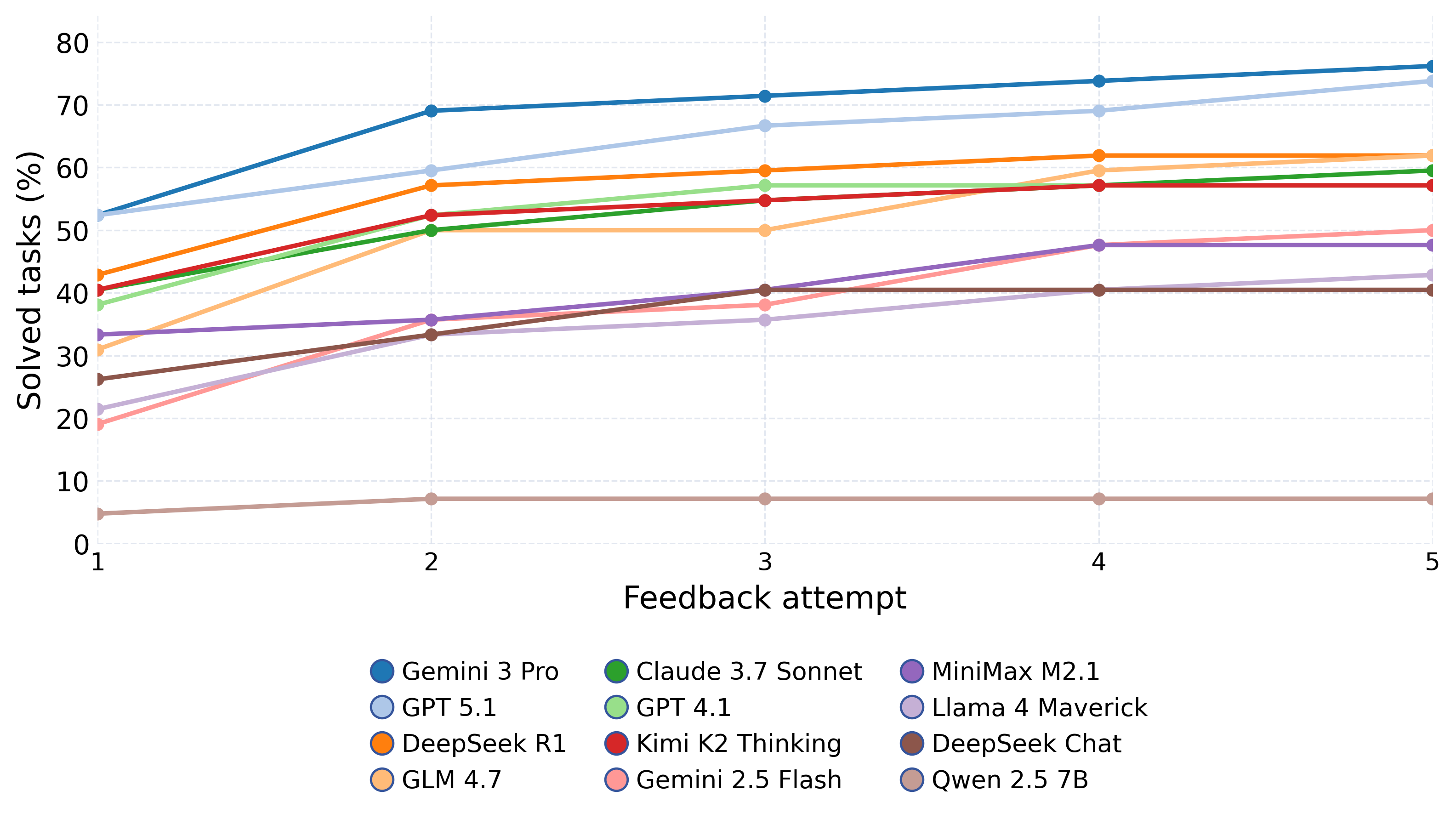}
\caption{\textbf{Cirq follows the same early-gain, late-plateau pattern.} Additional repair attempts help most in the first few rounds.}
\label{fig:app_pass1fb_cirq}
\end{figure}

\FloatBarrier
\noindent\textbf{Observations:} \textbf{(i)} Performance increases monotonically with additional feedback attempts, which confirms that iterative repair generally improves functional correctness. \textbf{(ii)} Most gains arrive early (attempts 1$\rightarrow$2), followed by diminishing returns after roughly three attempts. \textbf{(iii)} Qiskit (Fig.~\ref{fig:app_pass1fb_qiskit}) saturates earlier for the strongest models, whereas PennyLane (Fig.~\ref{fig:app_pass1fb_pennylane}) and Cirq (Fig.~\ref{fig:app_pass1fb_cirq}) often improve more gradually through attempts 4--5. \textbf{(iv)} Feedback compresses the spread among stronger models, but the weakest systems plateau quickly, which points to failure modes that retries do not resolve.

\FloatBarrier

\begin{figure}[htbp]
\centering
\includegraphics[width=\linewidth]{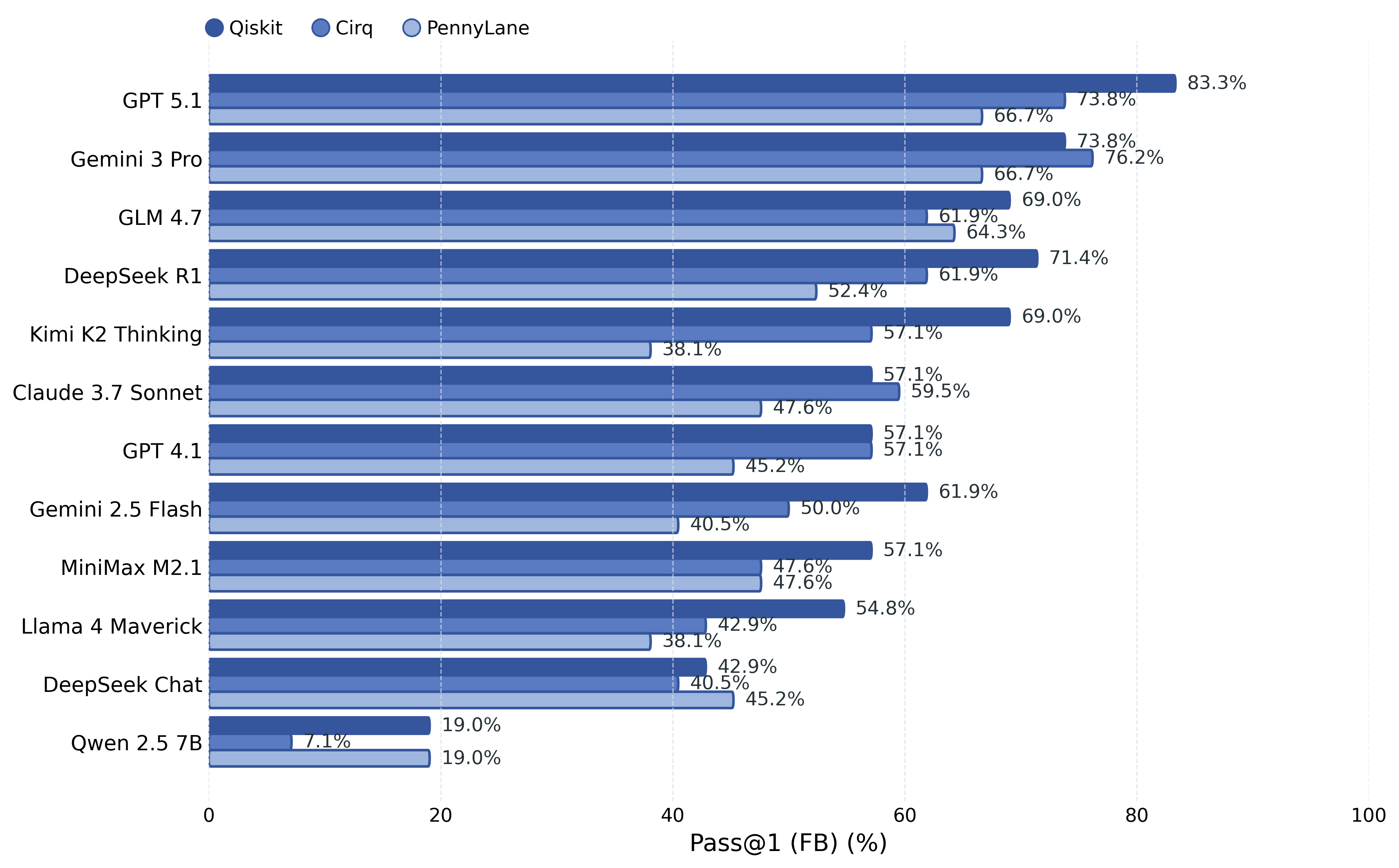}
\caption{\textbf{Feedback compresses the spread between models, but does not erase it.} Aggregate success rates after up to 5 repair attempts across all frameworks.}
\label{fig:app_pass1fb_all_frameworks}
\end{figure}

\begin{figure}[htbp]
\centering
\includegraphics[width=\linewidth]{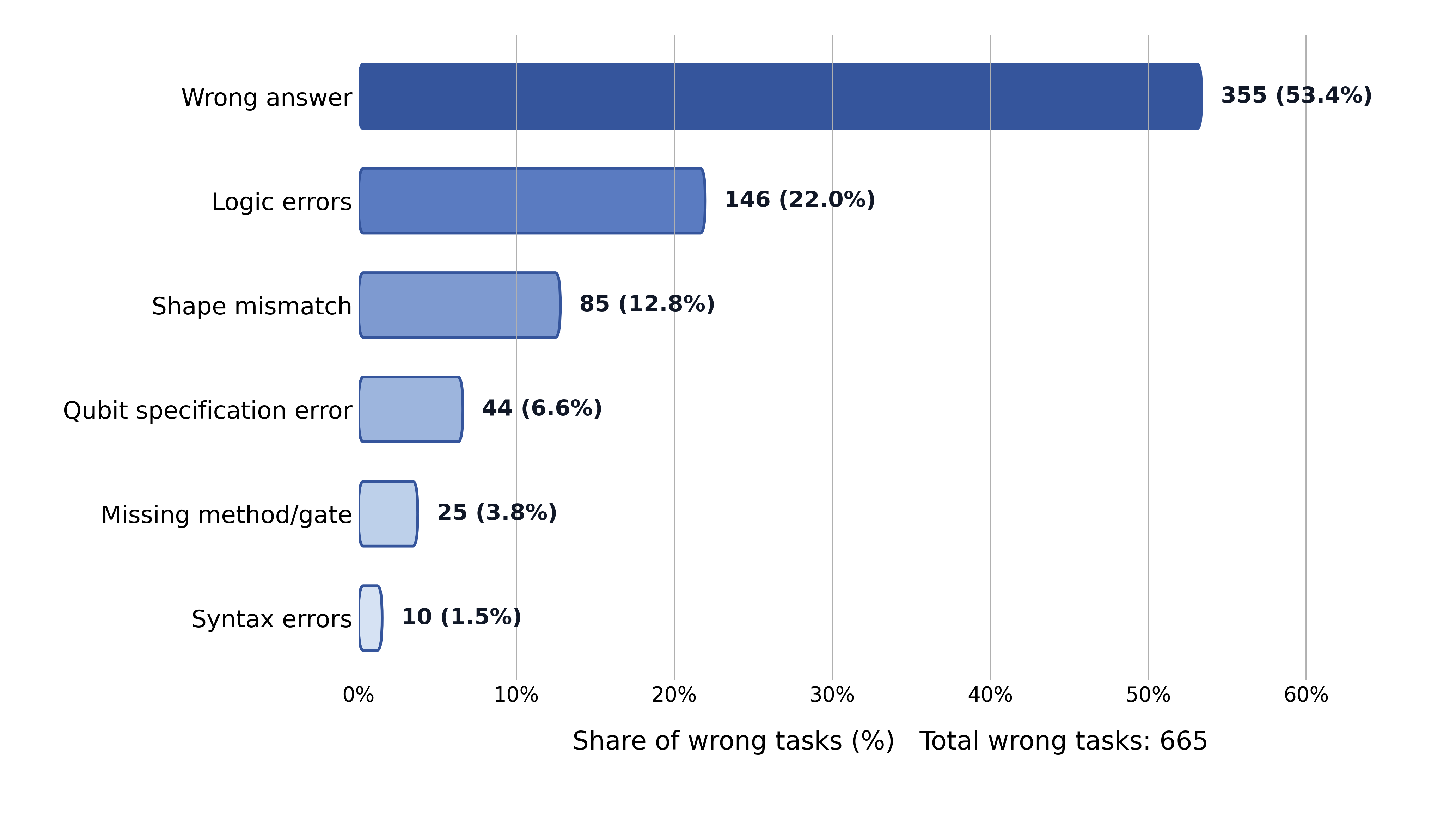}
\caption{\textbf{After repair, the remaining failures are mostly semantic.} Residual post-feedback errors become more concentrated in deeper reasoning mistakes.}
\label{fig:feedback_error_pie}
\end{figure}
\noindent\textbf{Observations:}
After the feedback loop, Fig.~\ref{fig:feedback_error_pie} shows that the total number of wrong tasks decreases substantially, from 977 to 665. The remaining errors are even more heavily concentrated in semantic issues, with wrong answers accounting for 53.4\% of failures, followed by logic errors (22.0\%) and shape mismatches (12.8\%). Surface-level implementation problems such as missing methods/gates (3.8\%) and syntax errors (1.5\%) become much less frequent. In other words, feedback is effective at fixing visible coding mistakes, but the harder remaining problem is still correct reasoning.